\documentclass[conference]{IEEEtran}
\usepackage{times}
\usepackage[T1]{fontenc}
\usepackage[utf8]{inputenc}

\usepackage[numbers,sort&compress]{natbib}
\usepackage{multicol}
\usepackage[pagebackref=true,breaklinks=true,bookmarks=true,colorlinks]{hyperref}

\usepackage{graphics} %
\usepackage{epsfig} %
\usepackage{times} %
\usepackage{amsmath} %
\usepackage{amssymb}  %
\usepackage{siunitx} %
\usepackage{textcomp}
\usepackage{gensymb} %
\usepackage{booktabs}
\usepackage{multirow}
\usepackage{xcolor}
\usepackage{bm}
\let\labelindent\relax
\usepackage{enumitem}
\usepackage{xspace}

\usepackage{makecell}
\usepackage{multicol}
\usepackage{rotating}
\usepackage[percent]{overpic}

\usepackage{contour}
\usepackage{courier}

\usepackage{subcaption} %
\usepackage[font=small]{caption}
\usepackage[percent]{overpic}

\usepackage[11pt]{moresize}

\usepackage{pifont}%

\usepackage[bookmarks=true]{hyperref}
\usepackage{xspace} 
\usepackage{xcolor} 
\usepackage{booktabs}       %
\usepackage{amsfonts}       %
\usepackage{nicefrac}       %
\usepackage{microtype}      %
\usepackage{amsmath,amsthm}
\usepackage{graphicx}
\usepackage{wrapfig}
\usepackage{listings}
\usepackage{algorithm}
\usepackage{xspace}
\usepackage{tabularx}
\usepackage{enumitem}
\usepackage{wrapfig}
\usepackage{microtype}
\usepackage{graphicx}
\usepackage{booktabs} %
\usepackage[noend]{algorithmic}
\definecolor{codegray}{rgb}{0.5,0.5,0.5}
\definecolor{codepurple}{rgb}{0.58,0,0.82}
\definecolor{backcolour}{rgb}{0.95,0.95,0.92}
\definecolor{commentcolour}{rgb}{0.0,0.6,0.0}
\definecolor{keywordcolour}{rgb}{0.0,0.0,0.6}
\definecolor{stringcolour}{rgb}{0.6,0.0,0.0}

\lstdefinestyle{mystyle}{
    backgroundcolor=\color{backcolour},   %
    commentstyle=\color{commentcolour}\ttfamily,
    keywordstyle=\color{keywordcolour}\bfseries,
    numberstyle=\tiny\color{codegray},
    stringstyle=\color{stringcolour},
    basicstyle=\fontfamily{\ttdefault}\scriptsize, %
    breakatwhitespace=false,     %
    breaklines=true,             %
    captionpos=b,                %
    keepspaces=true,             %
    numbersep=5pt,               %
    showspaces=false,            %
    showstringspaces=false,      %
    showtabs=false,              %
    tabsize=2,                   %
    frame=single,                %
    frameround=fttt,             %
    rulecolor=\color{codegray},  %
}

\lstdefinestyle{pythonstyle}{
  language=Python,                     %
  basicstyle=\ttfamily\small,          %
  keywordstyle=\color{keywordcolour},           %
  commentstyle=\color{commentcolour},            %
  showstringspaces=false,              %
  breaklines=true,                     %
  captionpos=b,                        %
  frame=single,                          %
  abovecaptionskip=1em,                %
  aboveskip=1em,                       %
  belowskip=1em,                       %
}

\usepackage[colorinlistoftodos]{todonotes}
\newcommand{\jdtext}[1]{\textcolor{black}{#1}}

\newcommand{\website}{\href{https://eureka-research.github.io/dr-eureka}{eureka-research.github.io/dr-eureka}}

\IEEEoverridecommandlockouts
\begin{document}

\title{DrEureka: \\ Language Model Guided Sim-To-Real Transfer\\[0.1cm]\large\website
\vspace{-0.5cm}}

\author{Yecheng Jason Ma$^{*1}$, William Liang$^{*1}$, Hung-Ju Wang$^{1}$, Sam Wang$^{1}$, \\Yuke Zhu$^{2,3}$, Linxi "Jim" Fan$^{2}$, Osbert Bastani$^{1}$, Dinesh Jayaraman$^{1}$
\\[1.0ex]
$^1$ University of Pennsylvania \quad\quad\quad $^2$ NVIDIA \quad\quad\quad $^3$ University of Texas, Austin
\thanks{$^*$Equal Contribution.}%
\\[-4.0ex]
}

\newcommand{\ourmethod}{\texttt{DrEureka}\xspace}
\newcommand{\jason}[1]{\textcolor{black}{#1}}
\newcommand{\jim}[1]{\textcolor{red}{Jim: #1}}
\newcommand{\yuke}[1]{\textcolor{blue}{Yuke: #1}}
\newcommand{\johnny}[1]{\textcolor{orange}{#1}}
\newcommand{\will}[1]{\textcolor{blue}{#1}}
\newcommand{\sam}[1]{\textcolor{black}{#1}}

\newcommand{\newtext}[1]{\textcolor{black}{#1}}
\newcommand{\arxiv}[1]{\textcolor{black}{#1}}

\maketitle

\begin{abstract}
Transferring policies learned in simulation to the real world is a promising strategy for acquiring robot skills at scale. However, sim-to-real approaches typically rely on manual design and tuning of the task reward function as well as the simulation physics parameters, rendering the process slow and human-labor intensive. In this paper, we investigate using Large Language Models (LLMs) to automate and accelerate sim-to-real design. Our LLM-guided sim-to-real approach, DrEureka, requires only the physics simulation for the target task and automatically constructs suitable reward functions and domain randomization distributions to support real-world transfer. We first demonstrate that our approach can discover sim-to-real configurations that are competitive with existing human-designed ones on quadruped locomotion and dexterous manipulation tasks. Then, we showcase that our approach is capable of solving novel robot tasks, such as quadruped balancing and walking atop a yoga ball, without iterative manual design.

\end{abstract}

\IEEEpeerreviewmaketitle

\begin{figure*}[htbp!]
    \centering
    \includegraphics[width=0.95\linewidth]{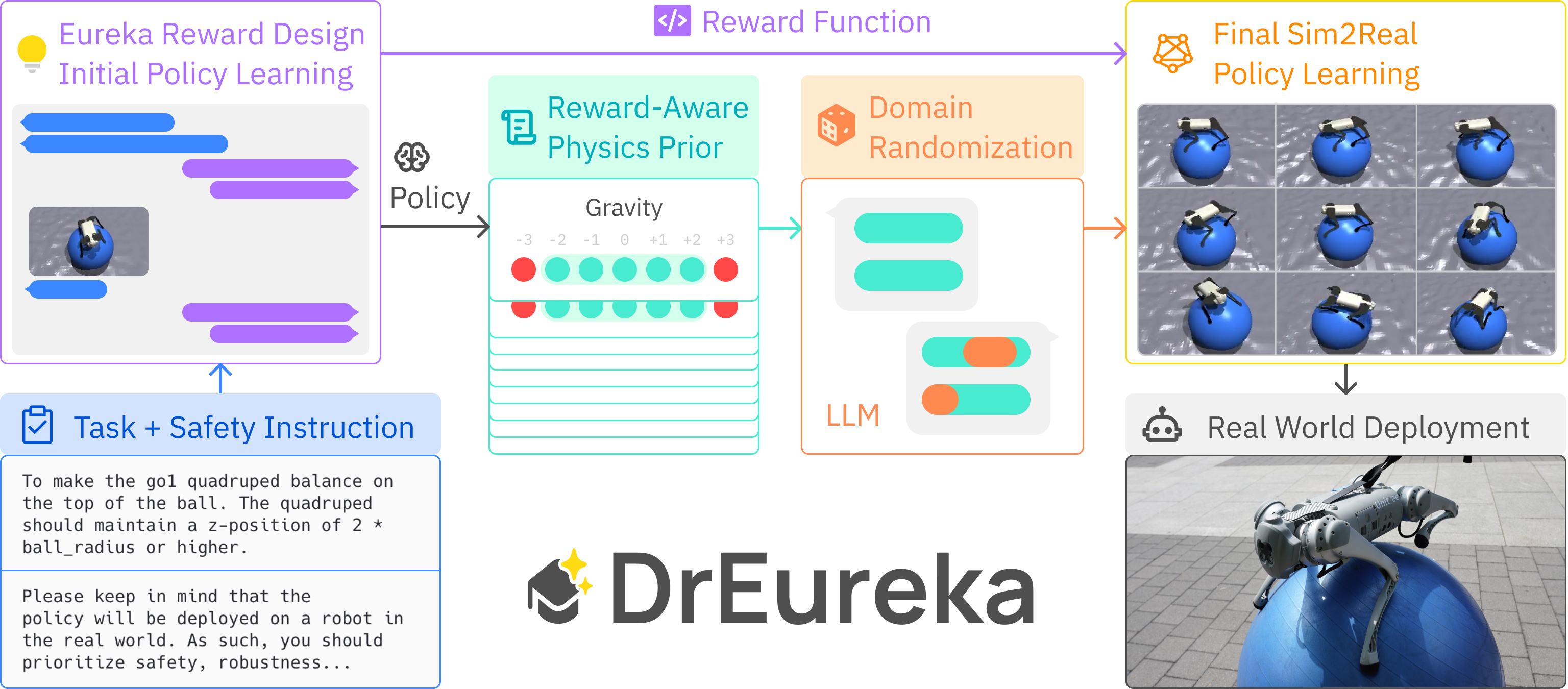}
    \caption{\ourmethod takes the task and safety instruction, along with environment source code, and runs Eureka to generate a regularized reward function and policy. Then, it tests the policy under different simulation conditions to build a reward-aware physics prior, which is provided to the LLM to generate a set of domain randomization (DR) parameters. Finally, using the synthesized reward and DR parameters, it trains policies for real-world deployment.}
    \label{fig:concept}
\end{figure*}

\section{Introduction}

Given their internet-scale training data, large language models (LLMs) have emerged as effective sources of common sense priors for robotics~\citep{ahn2022can,singh2023progprompt,huang2023grounded,liang2023code,brohan2023rt,kwon2023language}. Directly synthesizing robot policies from LLMs is difficult because it does not explicitly reason through the physics of the environment, however, when a simulator is available, we can combine the impressive world knowledge of LLMs together with the approximate physics knowledge in the simulator to learn complex low-level skills. Recent works have pursued this intersection and use LLMs to synthesize reward functions~\citep{yu2023language,xie2023text2reward,ma2023eureka} that can supervise robot reinforcement learning. However, thus far, these approaches have only been used in simulation, and transferring the policies to the real world still requires significant manual tuning of the simulators. In a typical process for sim-to-real policy synthesis, human engineers must manually and iteratively design reward functions and adjust simulation parameters until the configurations converge to enable stable policy learning. ~\citep{andrychowicz2020learning}. Thus, a natural question is whether we can additionally use LLMs to automate the components in the sim-to-real process that require intensive human efforts.

In this work, we propose \ourmethod (\textbf{D}omain \textbf{R}andomization Eureka), a novel algorithm that leverages LLMs to automate reward design and domain randomization parameter configuration simultaneously for sim-to-real transfer. While there are many sim-to-real techniques~\citep{aastrom1971system,jaquier2023transfer,muratore2022robot}, we focus on domain randomization because we believe that it is primed for LLMs to automate. Domain randomization (DR) is a family of approaches that apply randomization over a distribution of physical parameters in simulation, so that the learned policy can be robust against perturbance and transfers to the real world better~\citep{tobin2017domain,peng2018sim,muratore2022robot}. In DR, it is critical to select the right parameter distribution to ensure a successful transfer~\citep{vuong2019pick,kumar2021rma}. This step is often manually tuned by humans, because it is a challenging optimization problem that requires common sense physical reasoning (e.g., friction is important for walking on different surfaces) and knowledge of the robot system.
These characteristics of designing DR parameters make it an ideal problem for LLMs to tackle because of their strong grasp of physical knowledge~\citep{ahn2022can, wang2023newton} and effectiveness in \textit{generating \newtext{hypotheses}}, providing good initializations to complex search and black-box optimization problems in a zero-shot manner~\citep{yang2023large,zhang2023using,anonymous2024large,ma2023eureka,romera2023mathematical}. In \ourmethod, we show that these two distinct capabilities of LLMs can make them effective automated designers for DR configurations.

However, jointly optimizing for both reward functions and domain randomization parameters requires searching in a vast, infinite-dimensional function space, which is expensive and inefficient for LLMs to perform.
Instead, \ourmethod decomposes the optimization into three stages: an LLM first synthesizes reward functions, then an initial policy is rolled out in perturbed simulations to create a suitable sampling range for physics parameters, which is finally used by the LLM to generate valid domain randomization configurations. Specifically, to generate the highest quality of reward functions, we build on Eureka~\citep{ma2023eureka}, a state-of-the-art LLM-based reward design algorithm that can generate free-form, effective reward functions in code. To make Eureka reward functions more amenable for real-world transfer, we propose to include safety instructions in the prompt to automatically generate reward functions that induce safer behavior. Then, equipped with the best reward candidate as well as the associated policy, \ourmethod constructs reward-aware physics priors (RAPP) over environment physics parameters by evaluating the policy on various perturbed simulation dynamics; this procedure grounds the effective search ranges for LLM sampling of domain randomization configurations. Finally, the LLM receives the reward-aware prior as context and generates several DR distribution candidates to re-train policies more suitable for real-world deployment. Altogether, \ourmethod is a language-model driven pipeline for sim-to-real transfer without human intervention. A conceptual overview of the full algorithm is shown in Figure~\ref{fig:concept}.

 We evaluate \ourmethod on \newtext{quadruped and dexterous manipulator platforms, demonstrating that} our method is general and applicable to \newtext{diverse} robots and tasks. Our experiments focus on these two domains because reward design, domain randomization, and sim-to-real reinforcement learning at large have already established as critical components of effective policy learning strategies \newtext{within these domains}~\citep{rudin2022learning,lee2020learning,kumar2021rma,margolis2022rapid,margolis2023walk,akkaya2019solving,handa2023dextreme,qi2023hand}. Naturally, there are well-tested, open-sourced simulation environments that provide ideal testbeds for assessing \ourmethod's capability for supervising sim-to-real transfer~\citep{rudin2022learning,margolis2023walk,shaw2023leap}; as a reference point, our main comparison is \newtext{with two} existing human-designed \newtext{configurations}~\citep{margolis2022rapid,shaw2023leap} in order to demonstrate that \ourmethod can autonomously achieve useful sim-to-real designs. On \newtext{quadruped locomotion}, \ourmethod-trained policies outperform those trained with human-designed reward functions and DR parameters by 34\% in forward velocity and 20\% in distance travelled across various real-world evaluation terrains. \newtext{In dexterous manipulation, \ourmethod's best policy performs nearly 300\% more in-hand cube rotations than the human-developed policy within a fixed time period.} Through extensive ablation studies, we confirm that the components of DrEureka are all essential to generating effective safety-regularized reward functions. Finally, to demonstrate how \ourmethod can be used to accelerate sim-to-real on a new task, we test \ourmethod on the novel and challenging walking globe task commonly seen in circus, where the quadruped attempts to balance \newtext{and walk} on a yoga ball for as long as possible. Trained with \ourmethod, our policy can stay balanced on a real yoga ball for \newtext{minutes} on diverse indoor and outdoor terrains with minimal safety support.

In summary, our contributions are:
\begin{enumerate}
    \item \ourmethod, an LLM-guided sim-to-real algorithm that can automatically synthesize effective reward and domain randomization designs for sim-to-real transfer; 
    \item Extensive real world validation and analysis of \ourmethod on \newtext{representative robot tasks}; and
    \item Demonstration on a novel, challenging task.
\end{enumerate}

\begin{figure}
\label{fig:unitree}\includegraphics[width=\columnwidth]{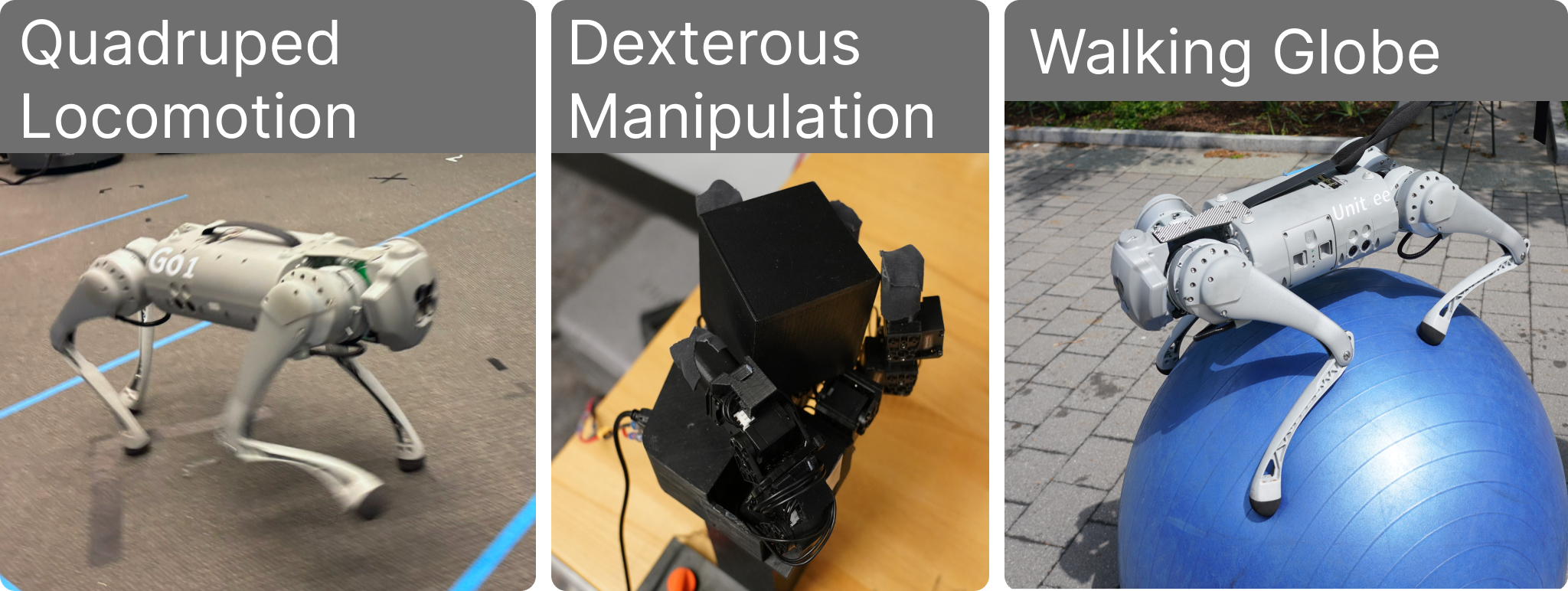}
\vspace{-0.4cm}
\caption{Our \newtext{quadruped locomotion, dexterous cube rotation, and walking globe} tasks. Walking globe is a novel task to show \ourmethod's capability for guiding the sim-to-real transfer of a challenging new task without pre-existing sim-to-real configurations.}
\label{figure:environments}
\vspace{-0.2in}
\end{figure}

\section{Related Work}

\textbf{Large Language Models for Robotics.}
Large Language Models (LLMs) have demonstrated capabilities as semantic planners~\citep{ahn2022can,huang2023grounded,singh2023progprompt,zhang2023bootstrap}, action models~\citep{brohan2023rt, szot2023large,tang2023saytap}, and symbolic programmers~\citep{liang2023code, singh2023progprompt, wang2023demo2code, huang2023instruct2act, wang2023voyager, liu2023llm+, silver2023generalized, ding2023task, lin2023text2motion, xie2023translating} for robotics applications. Recent works have explored using LLMs to guide the learning of low-level skills via reward function~\citep{yu2023language, xie2023text2reward, ma2023eureka} and environment design~\citep{wang2023gensim,wang2023robogen}; however, to the best of our knowledge, no prior work has explicitly studied whether LLMs can automate various design aspects of the sim-to-real procedure. In this work, we focus on the two important bottlenecks of reward design and domain randomization and introduce a novel technique that leverages LLMs' capability as solution generators for challenging optimization problems to automate sim-to-real transfer design.

\textbf{Domain Randomization.} 
To bridge the gap in physical dynamics between the real world and its simulation counterpart, domain randomization (DR) perturbs simulation physics parameters, such as friction and restitution, to improve the transferability of policies trained in simulation~\citep{tobin2017domain,peng2018sim,muratore2022robot}. The most common DR approach is to uniformly sample simulation parameters from a fixed distribution~\citep{tobin2017domain,peng2018sim,andrychowicz2020learning, qi2023hand, kumar2021rma}. To improve upon this simple randomization strategy, some works propose to automatically adjust the randomization distribution based on simulation training performance~\citep{akkaya2019solving, handa2023dextreme, tiboni2023domain}. Beyond feedback from simulation, some works have sought to use small amount of real-world policy trajectories to iteratively calibrate the randomization distributions in simulation to better adapt to the real world~\citep{mehta2020active, ramos2019bayessim,chebotar2019closing, muratore2021data}. Despite progress in DR algorithms, the form (i.e., which parameters to randomize) and the initial sampling distributions are typically manually chosen by practitioners with domain expertise, and these design choices have been shown to have large effect on the downstream policy performance~\citep{vuong2019pick,xie2021dynamics}. Our work is the first to study whether LLMs can automatically synthesize domain randomization configurations. 

\textbf{Sim-to-Real Robot Learning.} Beyond domain randomization, sim-to-real robot learning has been extensively studied in the literature with many complementary techniques, such as system identification~\citep{aastrom1971system,yu2017preparing,tan2018sim}, domain adaptation~\citep{pinto2017robust,nagabandi2018learning,bousmalis2018using,james2019sim,ren2023adaptsim}, transfer learning~\citep{jaquier2023transfer}, and many others~\citep{muratore2022robot}. These approaches differ from domain randomization in that they assume some interaction data with the real-world environment to bridge the sim-to-real gap. At the intersection of sim-to-real and LLMs, prior works have demonstrated policies synthesized in LLM-guided simulation training environments can transfer to the real world~\citep{ha2023scaling,wang2023gensim,yu2023language,xie2023text2reward}. However, aspects of the training pipelines pertaining to the sim-to-real transfer itself are still manually designed in these prior works. To the best of our knowledge, ours is the first work to investigate whether LLMs themselves can be used to guide sim-to-real transfer, and in particular, combining automated reward design and domain randomization for highly agile skills.

\section{Problem Setting}
We formalize the sim-to-real design problem setting. In a sim-to-real design instance, we assume a target real-world environment and a simulation environment without a built-in reward function or domain randomization configuration. The goal of sim-to-real RL is to train a policy in the simulation environment and then directly transfer it to the target real-world environment without further training. 

Mathematically, the simulation environment can be defined as a Markov Process $M=(S, A, T)$, in which $S$ is the environment state space, $A$ the action space, and $T$ the transition function. This assumption is easy to satisfy in practice, e.g., by porting the URDF files of the robot and object models into a simulator. For a task, we represent the true task objective with $F: \Pi \rightarrow \mathbb{R}$, which maps a policy to a numerical value that indicates its performance. A sim-to-real algorithm \texttt{Algo} for reward design and domain randomization takes $M$ and task specification $l_{\text{task}}$ as inputs, and outputs a reward function $R$ and a \textit{distribution} over transition functions,
 $\mathcal{T}$:
 \begin{equation}
     \mathcal{T}, R \leftarrow \texttt{Algo}(M, l_{\text{task}})
 \end{equation}
 Then, a policy learning algorithm $\mathcal{A}$ outputs a policy
\begin{equation}
\pi \leftarrow \mathcal{A}(M, \mathcal{T}, R),
\end{equation}
which is evaluated in the unknown true MDP (i.e., the real world environment) $M^*$
\begin{equation}
    f^* := F_{M^*}(\pi) 
\end{equation}
The goal of sim-to-real is to design $\mathcal{P}$ and $\mathcal{R}$ to maximize $f^*$:
\begin{equation}
    \label{eq:sim-to-real-objective}
    \max_{\mathcal{T}, \mathcal{R}} F_{M^*}(\mathcal{A}(M, \mathcal{T}, \mathcal{R}))
\end{equation}

Commonly, \texttt{Algo} is a human engineer who manually designs  $\mathcal{T}$ (i.e., domain randomization) and $\mathcal{R}$ (i.e., reward engineering). Specifically, the simulator comes with a set of physics parameters (e.g., mass and friction of objects) $\mathcal{P}$ whose values can be set and sampled according to a distribution. Domain randomization (DR) involves (1) selecting a set of physics parameters $\{p\} \subseteq \mathcal{P}$, and (2) selecting a randomization range for each of the chosen parameters. On the other hand, reward engineering tasks the human to write a dense reward function code for task $l$, and typically involves a trial-and-error procedure where the human observes policies trained using the current reward to design new reward candidates~\citep{russell1995artificial,sutton2018reinforcement,booth2023perils}.

In this work, we investigate whether LLMs, equipped with their physical common sense priors and solution generation capability, can guide and automate the sim-to-real design steps. That is, \texttt{Algo} is a LLM that ingests task specification $l_{\text{task}}$ in natural language and $M$ as a program, which is satisfied in practice as simulation environments are implemented in code. The LLM then outputs $\mathcal{T}$ and $R$ as strings, which are compiled into suitable programmatic formats for downstream policy learning.

\section{Method}
In this section, we introduce \ourmethod, which uses LLMs to automate two important bottlenecks in sim-to-real design: reward design and domain randomization. At a high level, \ourmethod first uses the LLM to generate a reward function that is both effective at the task and safe (Section~\ref{sec:reward} \& \ref{sec:safety}), then uses the resulting simulation policy to construct a prior distribution over randomizable parameters (Section~\ref{sec:prior}), and finally instructs the LLM to generate suitable domain randomization configurations based on the prior (Section~\ref{sec:domain-randomization}).

\subsection{Background: Eureka Reward Design}
\label{sec:reward}
Our reward design component builds on Eureka~\citep{ma2023eureka} due to its simplicity and expressivity but introduces several improvements to enhance its applicability for sim-to-real settings. In Eureka, the LLM first takes the task description $l_{\text{task}}$ and a summary of the environment state and action spaces (provided by environment code $M$) as input, and then samples several reward functions as code. Each reward function candidate is evaluated by training policies via reinforcement learning using that reward, and computing task scores $F$ for these policies. These scores as well as other training statistics (e.g., values of the reward components during training) are provided as feedback to the LLM to iteratively \textit{evolve} better reward functions that maximize $F$. The final output of Eureka is the best reward and policy pair 
\begin{equation}
    R_{\text{Eureka}}, \pi_{\text{Eureka}} := \texttt{Eureka}(M, l_{\text{task}}) 
\end{equation}

\subsection{Safety Instruction}
\label{sec:safety}
In Eureka, an implicit assumption is that the target environment $M^*$ \textit{is} the training simulation environment $M$. 
This is undesirable in the sim-to-real setting because a higher simulation score can often be achieved by over-exerting the robot motors or learning unnatural behavior, \sam{consequently encouraging} the LLM reward candidate sampler to favor reward functions that do not include safety terms (e.g., torque magnitude penalty). To mitigate this problem, one approach is post-hoc adding safety terms to $R_{\text{Eureka}}$. But this approach requires manually defining the safety terms and also fails to consider how the safety terms interact with other task-relevant components in $R_{\text{Eureka}}$. If the scale of the safety term dominates other terms, this approach may inadvertently induce degenerate behavior that is overly conservative~\citep{kim2023not}. 

Instead, we propose to directly exploit the strong instruction-following capability of instruction-tuned LLMs~\citep{ouyang2022training} and prompt the LLM to explicitly consider including safety terms for stability, smoothness, and desirable task-specific attributes as a part of the language specification $l$: 
\begin{equation}
\label{eq:eureka-safety}
    R_{\text{DrEureka}}, \pi_{\text{initial}} := \texttt{Eureka}(M, l_{\text{task}}+l_{\text{safety}}) 
\end{equation}
We hypothesize that this allows the LLM to naturally balance the weighting and potentially non-additive interactions of all reward components, thereby enabling better real-world transfer. See Algorithm~\ref{algo:eureka} for pseudocode.

\begin{algorithm}[]
\caption{\ourmethod Reward Design}
\label{algo:eureka}
\begin{algorithmic}[1]
\small
\STATE \textbf{Require}: Task description $l_{\text{task}}$, safety instruction $l_{\text{safety}}$, \\ RL algorithm $\mathcal{A}$, environment code $M$, 
coding LLM $\texttt{LLM}$, fitness function $F$, initial prompt $\texttt{prompt}$
\STATE \textbf{Hyperparameters}: search iteration $N$, iteration batch size $K$
\FOR{\text{N iterations}}
\STATE \textcolor{purple}{\texttt{// Sample reward functions from LLM}}
\STATE $R_1,..., R_k \sim \texttt{LLM}(l_{\text{task}}::l_{\text{safety}}, M, \texttt{prompt})$ 
\STATE \textcolor{purple}{\texttt{// Train policies in simulation}}
\STATE $\pi_1 = \mathcal{A}(M, R_1),..., \pi_k = \mathcal{A}(M, R_k)$
\STATE \textcolor{purple}{\texttt{// Evaluate policies \jdtext{in simulation}}}
\STATE $s_1 =F(\pi_1), ..., s_K = F(\pi_k)$
\STATE \textcolor{purple}{\texttt{// Reward reflection}}
\STATE $\texttt{prompt} := \texttt{prompt} :: \texttt{Reflection}(R^n_{best}, s^n_{best})$,\\ 
where $best = \arg \max_{k} s_1,...,s_K$
\STATE \textcolor{purple}{\texttt{// Update best reward and policy}}
\STATE $R_{\text{DrEureka}}, \pi_{\text{initial}}, s_{\text{DrEureka}} = (R^n_{\text{best}}, \pi^n_{\text{best}} s^n_{\text{best}})$, \quad if $s^n_{\text{best}} > s_{\text{Eureka}}$
\ENDFOR
\STATE \textbf{Output}:  $R_{\text{DrEureka}}, \pi_{\text{initial}}$
\end{algorithmic}
\end{algorithm}

\subsection{Reward-Aware Physics Prior}
\label{sec:prior}

A safe reward function can regularize the policy behavior fixing a choice of environment, but is not in itself sufficient for sim-to-real transfer. Given $R_{\text{DrEureka}}$ and $\pi_{\text{initial}}$, how should we prompt the LLM to generate effective domain randomization configurations? This is a challenging problem because we do not have access to the real-world environment $M^*$ at training time. However, we do have access to $M$, which comes with default values for simulation physics parameters. 
\newtext{Even so}, the default values themselves are not sufficient as guidance for the LLM because they reveal no information about the parameter scales and base ranges to sample. Simulation physical parameters often have built-in ranges (i.e., max and min values), but we hypothesize and validate that these ranges are too wide and may significantly hamper policy learning~\citep{vuong2019pick,andrychowicz2020learning}. 

We introduce a simple \textit{reward aware physics prior (RAPP)} mechanism to restrict the base ranges for the LLM. \jason{At \newtext{a} high level, RAPP seeks for the maximally diverse range of environment parameters that $\pi_{\text{initial}}$ is still performant.} Our insight is that domain randomization should be dependent on the task reward function and customized to the policy behavior learned without domain randomization.
For instance, randomizing frictions over too wide of a range is likely to sample friction values that are infeasible to learn given the reward function. 
In practice, for each domain randomization parameter, RAPP computes a lower and upper bound of values that are ``feasible'' for training. Specifically, we search through a general range of potential values at varying magnitudes, and with each value, we set it in simulation (keeping all other parameters at default) and roll out $\pi_{\text{Eureka}}$ in this modified simulation. If the policy's performance satisfies a pre-defined success criterion, we deem this value as feasible for this parameter. Given the set of all feasible values for each parameter, our lower and upper bounds for a parameter are the minimum and maximum feasible values. It is computationally light since it requires only evaluating the policies under different physics parameters and can be efficiently done in parallel.

\begin{algorithm}[H]
\caption{Reward Aware Physics Prior (RAPP)}
\label{algo:physics-prior}
\begin{algorithmic}[1]
\small
\STATE \textbf{Require}: Reinforcement learning policy $\pi_{\text{initial}}$, simulator $S$, success criteria $F$, domain randomization parameters $\mathcal{P}$ and their respective search values $\mathcal{R}$, 
\FOR{\text{randomization parameter $p \in \mathcal{P}$}}
\STATE \textcolor{purple}{\texttt{// Initialize output range to extremes}}
\STATE $l = \inf, h = -\inf$
\FOR{\text{search value $r \in \mathcal{R}$}}
\STATE \textcolor{purple}{\texttt{// Change one randomization parameter while leaving others at default value}}
\STATE $S.p = r$
\STATE \textcolor{purple}{\texttt{// Evaluate policy in simulation, record trajectory $\tau$}}
\STATE $\tau = S(\pi_{\text{initial}})$
\STATE \textcolor{purple}{\texttt{// Evaluate success criteria, update range if successful}}
\IF{$F(\tau)$}
\STATE $l = \min(l, r)$
\STATE $h = \max(h, r)$
\ENDIF
\ENDFOR
\ENDFOR
\STATE \textbf{Output}: $l, h$ for each $p \in \mathcal{P}$
\end{algorithmic}
\end{algorithm}

\subsection{LLM for Domain Randomization}
\label{sec:domain-randomization}

\begin{figure}
\includegraphics[width=\columnwidth]{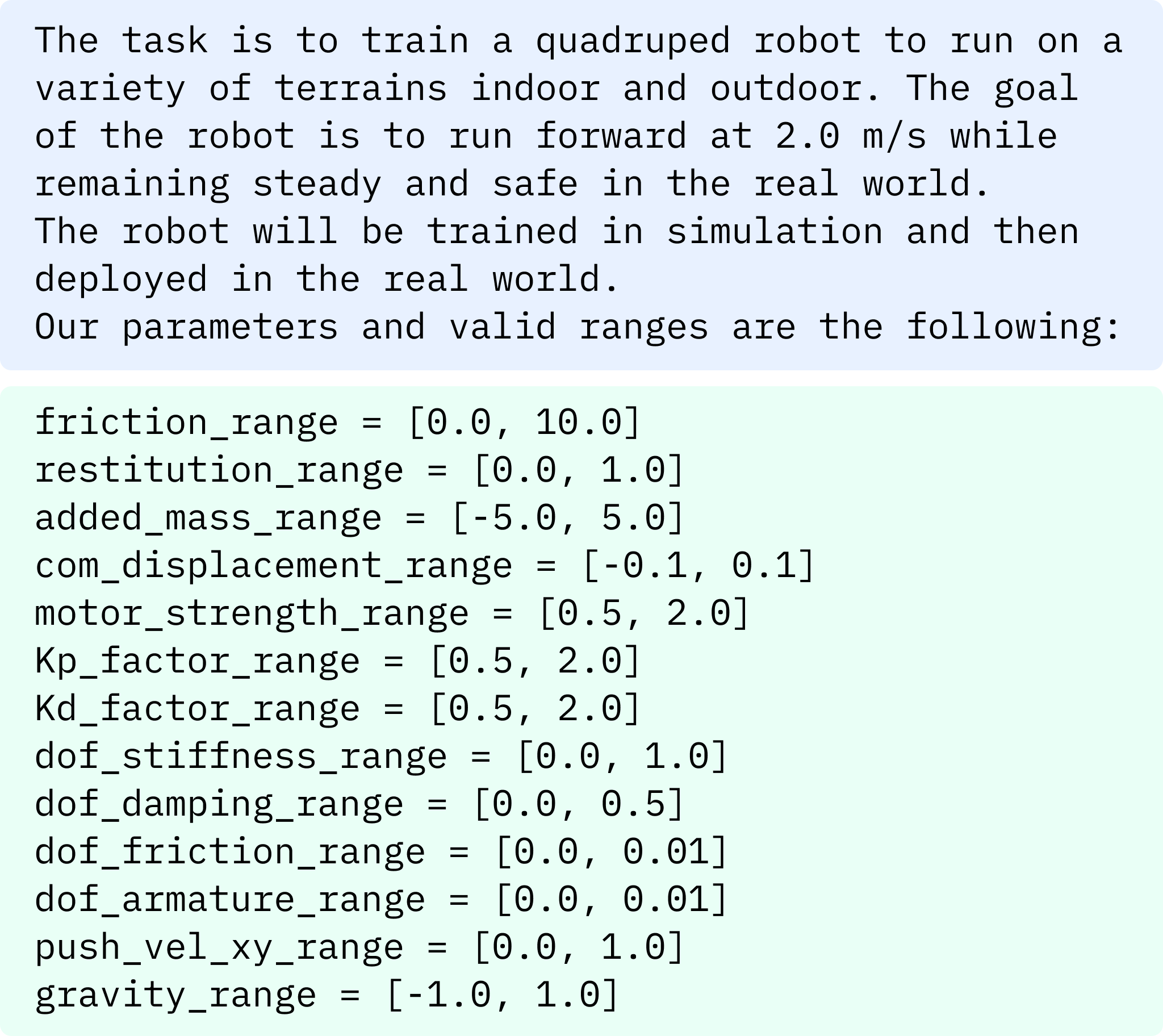}
\caption{\ourmethod prompt for generating domain randomization parameters. The blue paragraph describes the instruction, and the green paragraph is the reward aware parameter prior computed in Algorithm~\ref{algo:physics-prior}.}
\label{figure:dr-prompt-example}
\end{figure}

Given the RAPP ranges for each DR parameter, the final step of \ourmethod instructs the LLM to generate domain randomization configurations within the limits of the RAPP ranges. Compare this to automatic domain randomization~\citep{akkaya2019solving,handa2023dextreme}: they too search for parameter ranges where the policy performs well, but they directly set the DR parameters to this range. Instead, we use it as a guide for LLM. Our experiments show that this performs better as the base range can be too wide and hampers policy learning. Concretely, we provide all randomizable parameters $\mathcal{P}$ and their RAPP ranges in the LLM context and ask the LLM (1) to choose a subset of $\{p\} \subseteq \mathcal{P}$ to randomize and (2) determine their randomization ranges. See Figure~\ref{figure:dr-prompt-example} for the actual domain randomization prompt \ourmethod uses in our experiments. In this manner, the backbone LLM zero-shot generates several independent DR configuration samples, $\mathcal{T}_1,..., \mathcal{T}_m$. Finally, we use RL to train policies for each reward and DR combination, resulting in a set of policies where 

\begin{equation}
    \pi_{\text{final,i}} = \mathcal{A}(M, \mathcal{T}_i, R_{\text{DrEureka}}), i=1,...m
\end{equation}

Unlike \sam{reward design}, it is difficult to select the \textit{best} DR configuration and policy in simulation as each policy is trained on its own DR distribution and cannot be easily compared. Hence, we keep all $m$ policies and report both the best and the average performance in the real world. Finally, note that some prior works prescribe \sam{continuously tuning} the DR configuration to adapt to improving policy capabilities over the course of training~\citep{ramos2019bayessim,akkaya2019solving,handa2023dextreme,tiboni2023domain}; we find in practice that the initial DR configurations generated by \ourmethod suffice for sim-to-real transfer in our setups without intra-training adaptation.

\footnotetext[1]{In both \textit{Without Prior} and \textit{Uninformative Prior} experiments, 15 out of the 16 policies resulted in jerky and dangerous behavior, many times immediately triggering the controller's motor power protection fault. We count these trials as 0m/s, 0m traveled.}
\section{Experimental Setup}

\textbf{Robots and Tasks.} We adopt commercially available, low-cost robots with well-supported open-sourced simulators as our evaluation platforms. For our main experiments on quadrupedal locomotion, we use Unitree Go1. The Go1 is a small quadrupedal robot with 12 degrees of freedom across four legs. Its observations include joint positions, joint velocities, and a gravity vector in the robot's local frame, as well as a history of past observations and actions. We use the simulation environment as well as the real-world controller from~\citet{margolis2022rapid}.
The task of forward locomotion is to walk forward at 2 meters-per-second on flat terrains; while it is possible for the robot to walk forward at a higher speed, we find 2 m/s to strike a good balance between task difficulty and safety as our goal is not to achieve the highest speed possible on the robot. In the real world, we set up a 5-meter track in the lab (see Figure~\ref{figure:robustness}) and measure the forward projected velocity and total meters traveled in the track direction.

\newtext{
In addition to locomotion, we validate \ourmethod's applicability to a second task category of dexterous manipulation. Here, we use the LEAP hand~\citep{shaw2023leap}, which is a low-cost anthropomorphic robot hand, featuring 16 degrees of freedom distributed among three fingers and a thumb. The task involves rotating a cube in-hand as many times as possible within a 20-second interval. This task is challenging because the policy only receives 16 joint angles and proprioceptive history, encoded via GRU~\citep{chung2014empirical}, as observations and does not have access to the position and the pose of the cube. The policy then outputs target joint angles as position commands to the motors.}

\newtext{
Both robots cost less than 10K USD and admit simulators in NVIDIA Isaac Gym~\citep{makoviychuk2021isaac} with sim-to-real training code that has been tested in the real world.
}

\textbf{Methods.} \ourmethod uses GPT-4~\citep{openai2023gpt4} as the backbone LLM\newtext{, and we use} the original Eureka hyperparameters for reward generation before sampling 16 DR configurations. To understand the best and the average performance of \ourmethod, we train policies for all 16 configurations and evaluate all policies in the real world. We primarily compare to \newtext{the} human-designed reward function and DR configuration from the original task implementations~\citep{margolis2022rapid, shaw2023leap} as reference;  We refer to this baseline as \texttt{Human-Designed}. Note that this baseline \newtext{for forward locomotion} trains a velocity-conditioned policy and utilizes a reward function with a velocity curriculum that gradually increases as policy training progresses. For our comparison, we train on the whole curriculum but evaluate the policy at 2 m/s. \newtext{We emphasize} that the purpose of comparing to \texttt{Human-Designed} is to determine whether \ourmethod can be \textit{useful} -- i.e., enabling sim-to-real transfer on a representative robot task for which robotics researchers have devoted time to designing effective sim-to-real pipelines. The absolute performance ordering is of less importance as LLMs and humans arrive at their respective sim-to-real configurations using vastly different computational and cognitive mechanisms.

To verify that a policy outputted by a reward-design algorithm itself is not effective for real-world deployment, we also compare against Eureka~\citep{ma2023eureka}, which designs rewards using LLMs without safety consideration and trains policies without domain randomization. 

\newtext{Additionally, we consider two classes of \ourmethod ablations that probe (1) whether some fixed DR configuration can generally outperform \ourmethod samples, and (2) the importance of \ourmethod's reward-aware priors (Section~\ref{sec:prior}) and LLM sampling (Section~\ref{sec:domain-randomization}). In the first class, we first compare to an ablation that does not train with domain randomization (\textbf{No DR}). Second, we consider a baseline that trains with the \texttt{human-designed} DR (\textbf{Human-Designed DR}) in the original implementation. Third, we consider a baseline that directly uses the full ranges of the RAPP parameter priors as the DR configuration (\textbf{Prompt DR}); this ablation can \newtext{be} viewed as applying domain randomization algorithms~\citep{akkaya2019solving,handa2023dextreme,tiboni2023domain} that seek to prescribe the maximally diverse parameter ranges where the policy performs well as the configurations. In the second category of ablations, we consider an ablation that only has access to the set of physics parameters but without the reward-aware priors (\textbf{No Prior}). Additionally, we consider an ablation that has only the default search range for RAPP as the parameter priors (\textbf{Uninformative Prior}). Finally, we consider a baseline that randomly samples from the RAPP ranges (\textbf{\newtext{Random Sampling}}); this baseline helps show whether LLM-based sampling is a better hypothesis generator. In all ablations, we fix the \ourmethod reward function for the task and only modify the DR configurations.}

\newtext{
Finally, we compare \ourmethod's DR-generation with prior methods based on Cross Entropy Method (CEM) ~\citep{vuong2019pick, KROESE201319, CrossEntropy} and Bayesian Optimization (BayRn) ~\citep{muratore2021data, frazier2018tutorial}, which optimize DR parameters by repeatedly training and evaluating policies in real. Note that while CEM and BayRn tackle the same problem, their iterative procedure is conceptually different from \ourmethod, which trains all policies in parallel; thus, this comparison favors the baselines because they use additional information from intermediate real-world evaluations. First, we consider CEM initialized with mean at simulation default values and variance 1 (\texttt{CEM Random}), following ~\citep{vuong2019pick}. Second, we consider CEM initialized by randomly sampling within the RAPP bounds, (\texttt{CEM RAPP}), which provides a stronger prior. Third, we consider BayRn with parameters bounded by RAPP and initial samples randomly drawn from RAPP (\texttt{BayRn RAPP}). Additional details are in the Appendix.
}

\textbf{Policy Training and Evaluation.} We train all policies entirely in the simulation environment and use policy training code framework provided by \citet{margolis2022rapid} \newtext{for forward locomotion and \citet{shaw2023leap} for cube rotation}. \newtext{For both tasks,} the reinforcement learning algorithm is Proximal Policy Optimization (PPO)~\citep{schulman2017proximal}. \newtext{Forward locomotion specifically uses a teacher-student variant of PPO} in which the teacher policy receives privileged state information in simulation to supervise a student policy that uses sensors available in the real world. Adopting the evaluation protocol from~\citet{ma2023eureka}, we use the original policy training hyperparameters for all policy training and do not modify or tune them for \ourmethod's configurations. Therefore, the differences in performance between \ourmethod and \texttt{Human-Designed} can be attributed to the different DR parameters as well as reward functions \ourmethod produces. For every DR configuration, we train policies using 3 random seeds and report average as well as standard deviation across trials and seeds. Video results are included \newtext{on our project website}.

\section{Results and Analysis}

\begin{table}
\centering
\resizebox{1\linewidth}{!}{
 \begin{tabular}{l|cc}
 \toprule
 {Sim-to-real Configuration} & Forward Velocity (m/s) & Meters Traveled (m) \\
 \midrule
 \texttt{Human-Designed}~\citep{margolis2022rapid} & 1.32 $\pm$ 0.44 & 4.17 $\pm$ 1.57  \\
  Eureka~\citep{ma2023eureka} & 0.0 $\pm$ 0.00 & 0.00 $\pm$ 0.00 \\ 
  Our Method (Best) & \textbf{1.83} $\pm$ 0.07 & \textbf{5.00} $\pm$ 0.00 \\ 
  Our Method (Average) & 1.66 $\pm$ 0.25 & 4.64 $\pm$ 0.78 \\ 
  \midrule 
  \multicolumn{3}{c}{Ablations for Our Method} \\
  \midrule
  Without DR & 1.21 $\pm$ 0.39 & 4.17 $\pm$ 1.04 \\
  With \texttt{Human-Designed} DR & 1.35 $\pm$ 0.16 & 4.83 $\pm$ 0.29 \\
  With Prompt DR & 1.43 $\pm$ 0.45 & 4.33 $\pm$ 0.58 \\
  Without Prior & 0.09 $\pm$ 0.36\footnotemark[1] & 0.31 $\pm$ 1.25 \\
  With Uninformative Prior & 0.08 $\pm$ 0.33\footnotemark[1] & 0.28 $\pm$ 1.13 \\
  With Random Sampling & 0.98 $\pm$ 0.45 & 2.81 $\pm$ 1.80 \\
  \midrule 
  \multicolumn{3}{c}{\newtext{DR Generation Baselines}} \\
  \midrule
  \newtext{\texttt{CEM Random}} & 0.00 $\pm$ 0.00 & 0.00 $\pm$ 0.00 \\
  \newtext{\texttt{CEM RAPP}} & 1.46 $\pm$ 0.12 & 5.00 $\pm$ 0.00 \\
  \newtext{\texttt{BayRn RAPP}} & 1.28 $\pm$ 0.62 & 4.00 $\pm$ 1.73 \\
  \bottomrule
 \end{tabular}}

   \caption{\textbf{Main comparison against baselines and ablations \newtext{for forward locomotion}.} \ourmethod's average and best policies outperform \texttt{Human-Designed} and a prior reward-design baseline. Ablations of the DR formulation in \ourmethod \newtext{and alternative baselines} all result in decreased performance.}
  \label{table:main-results}
\end{table}

\begin{table}
\centering
\resizebox{1\linewidth}{!}{
  \begin{tabular}{l|cc}
  \toprule
  {Sim-to-real Configuration} & Rotation (rad) & Time-to-Fall (s) \\
  \midrule
  \newtext{\texttt{Human-Designed}}~\citep{margolis2022rapid} & 3.24 $\pm$ 1.66 & 20.00 $\pm$ 0.00 \\ 
  \newtext{Our Method (Best)} & \textbf{9.39} $\pm$ 4.15 & \textbf{20.00} $\pm$ 0.00 \\ 
  \newtext{Our Method (Average)} & 4.67 $\pm$ 3.55 & 16.29 $\pm$ 6.28 \\ 
  \midrule 
  \bottomrule
 \end{tabular}}
  \caption{\newtext{\textbf{Comparison against \texttt{Human-Designed} for cube rotation.} Both the average and the best policies of \ourmethod surpass \texttt{Human Designed} in terms of total rotation.}}
  \label{table:leaphand-results}
\end{table}

Our experiments are designed to answer the following:
\begin{enumerate}
    \item Can \ourmethod be competitive with manual, pre-existing Sim2Real pipeline on known tasks?
    \item How important is each component of \ourmethod?
    \item Can \ourmethod help solve challenging new tasks for which no prior sim-to-real pipeline exists? 
\end{enumerate}

\subsection{Comparison to Pre-Existing Sim-to-Real Configurations}
We first directly compare \ourmethod to \texttt{Human-Designed} to assess whether \ourmethod is capable of providing sim-to-real training configurations comparable to human-designed ones. \newtext{For forward locomotion}, as shown in Table~\ref{table:main-results}, \ourmethod is able to outperform \texttt{Human-Designed} in terms of both forward velocity as well as distance traveled on the track. The \newtext{performance} of \ourmethod is robust across its different DR sample outputs; the average performance does not lag too far behind the best \ourmethod configuration and still \jdtext{performs on par with or slightly better than} \texttt{Human-Designed}. In contrast, the \jdtext{plain} Eureka generated policy fails to walk in the real world, validating that a reward design algorithm suitable for simulation is not sufficient for sim-to-real transfer. More details about our experiments comparing DrEureka’s reward against ablations can be found in the Appendix.

\newtext{
Similarly, for cube rotation, we see in Table~\ref{table:leaphand-results} that \ourmethod outperforms \texttt{Human-Designed} in terms of rotation while maintaining a competitive time-to-fall duration. We note that this task permits very little room for error; thus, policies generally perform very well or very badly, \sam{resulting} in the relatively larger standard deviation across \ourmethod's policies. Nevertheless, the best policy from \ourmethod significantly outperforms the baseline by nearly three times the rotation without dropping the cube. These results highlight the effectiveness and versatility of our approach across diverse robotic platforms.
}
        
\begin{figure}
\centering
\includegraphics[width=\columnwidth]{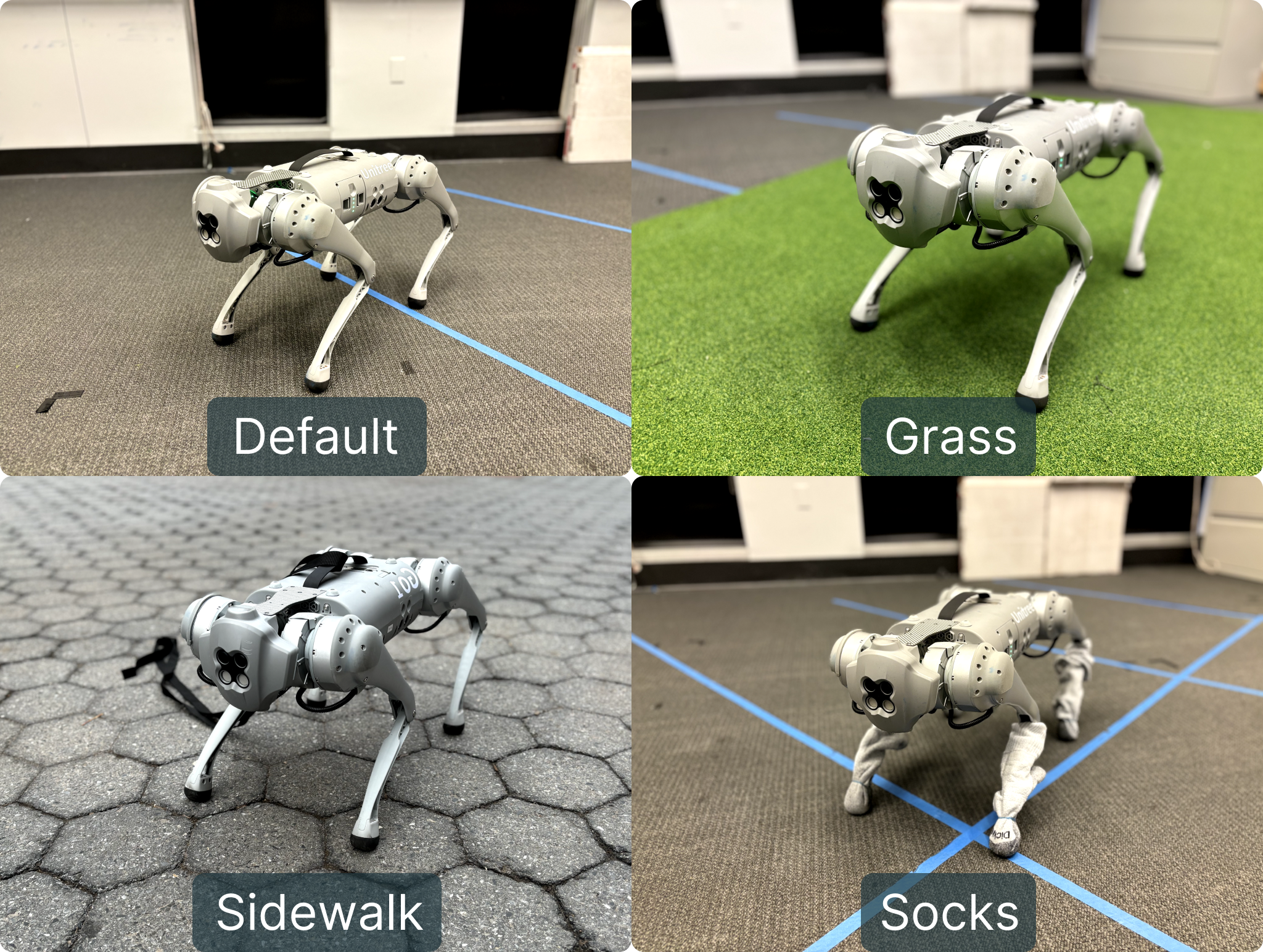}
\caption{The default real-world environment as well as additional environments to test \ourmethod's robustness for quadrupedal locomotion.}
\label{figure:robustness}
\end{figure}

\begin{figure}
\centering
\includegraphics[width=\columnwidth]{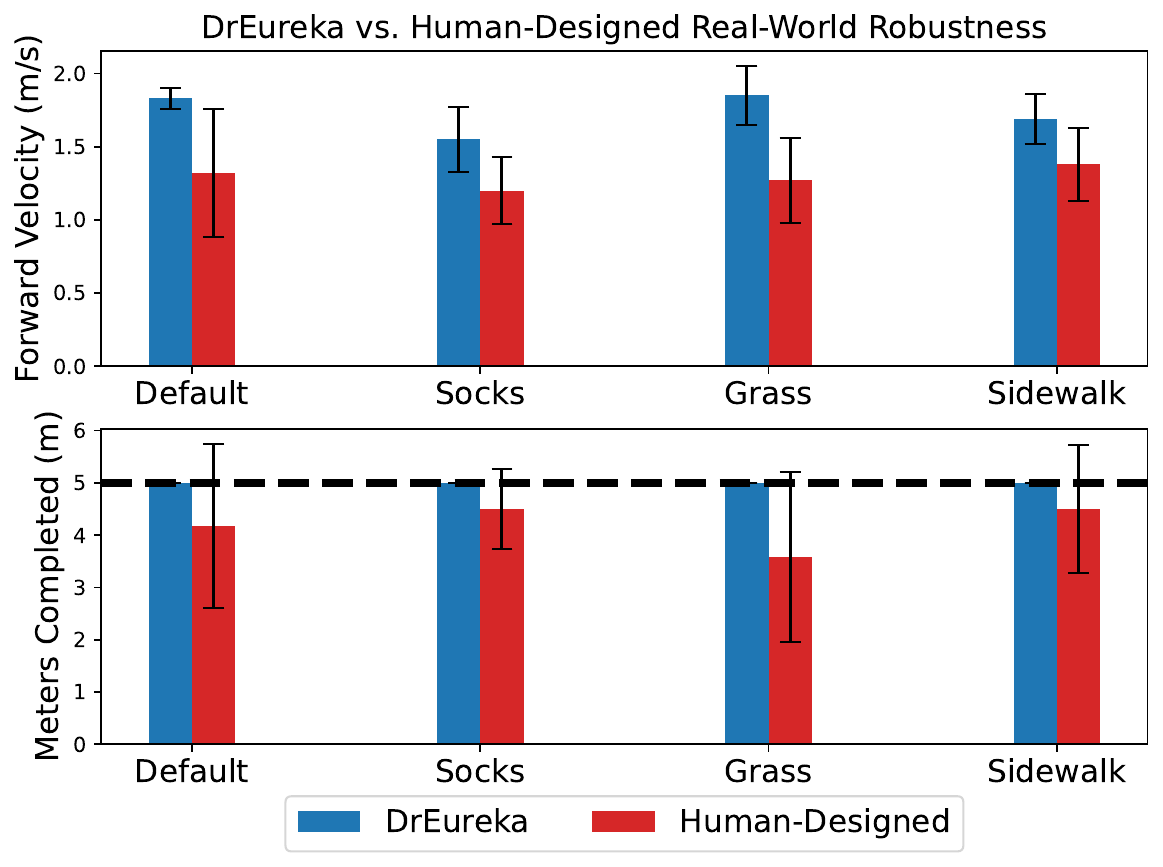}
\caption{\textbf{Real-world robustness evaluation.} \ourmethod performs consistently across different terrains and maintains advantages over \texttt{Human-Designed}.}
\label{figure:robustness-results}
\end{figure}

\textbf{Real-world robustness.} One main appeal of domain randomization is the robustness of the learned policies to real-world environment perturbations. To probe whether \ourmethod policies exhibit this capability, we test \ourmethod (Best) and \texttt{Human-Designed} on several additional testing environments \newtext{for forward locomotion} (Figure~\ref{figure:robustness}). Within the lab environment, we consider an artificial grass turf as well as putting socks on the quadruped legs. For an outdoor environment, we test on an empty pedestrian sidewalk; the results are shown in Figure~\ref{figure:robustness-results}. We see that across different testing conditions, \ourmethod remains performant and consistently matches or outperforms \texttt{Human-Designed}. This validates that \ourmethod is capable of producing robust policies in the real world. 

Having validated that \ourmethod can be as effective as human-crafted sim-to-real designs in real-world scenarios, we provide further analysis and perform ablations on the quadrupedal locomotion task to better understand the sources of its effectiveness.

\begin{figure*}
\centering
\includegraphics[width=\textwidth]{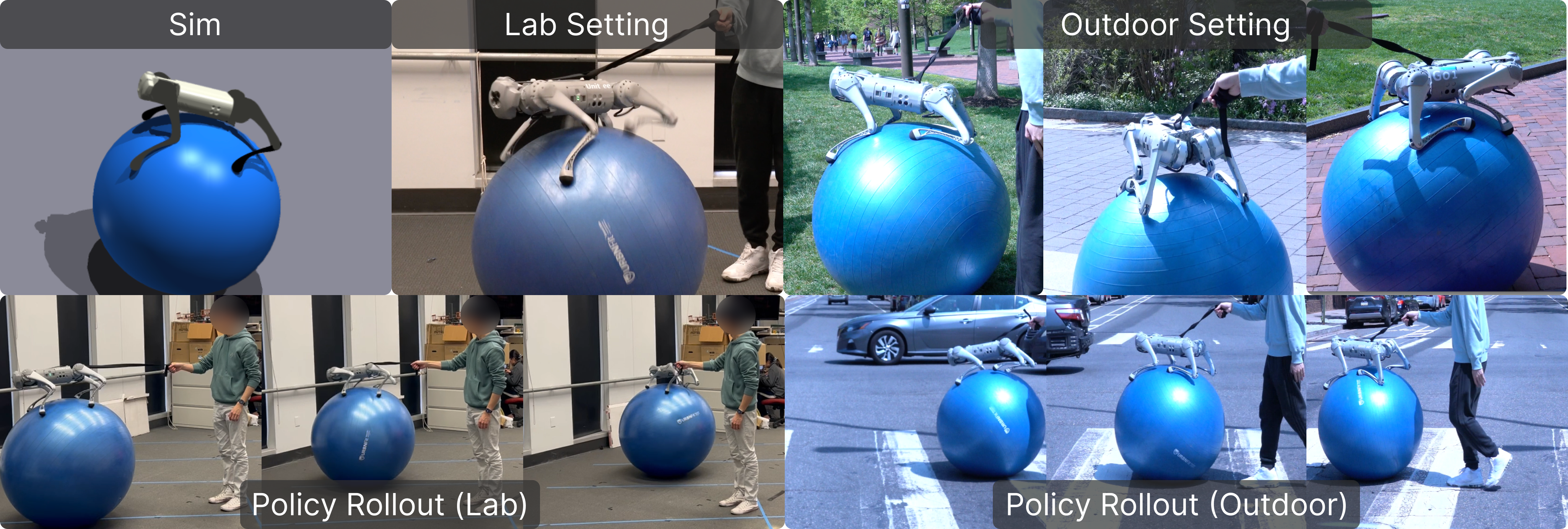}
\caption{Walking Globe sim and real environments. In lab settings, we loosely strap the robot horizontally to a center point to prevent robot from falling. For outdoor tests, we evaluate the policy across various terrains, including sidewalks, roads, grass, and wooden bridges.}
\label{figure:yoga}
\end{figure*}

\subsection{Does \ourmethod generate effective DR configurations?}
\label{sec:dr-ablation}

\newtext{
\textbf{\ourmethod surpasses DR optimization baselines.} In Table ~\ref{table:main-results}, we see that \texttt{BayRn RAPP} and \texttt{CEM RAPP} produce worse final policies and are less sample efficient than \ourmethod; this is despite the fact that several random initial samples from the RAPP range used to seed BayRn and CEM are reasonably effective. \texttt{CEM Random} performs worse as well since it has no access to the RAPP bounds, causing its variance of 1 to \sam{both overshoot and undershoot parameters}. These results show that LLM-generated DR can outperform prior feedback-based approaches. Moreover, since these baselines are iterative, \ourmethod incurs a significantly shorter wall-clock runtime of 3 hours, compared to 10 and 20 for CEM and BayRn. Finally, we note that \ourmethod avoids testing intermediate real-world policies that can be unsafe, especially for novel tasks like our globe walking task which lack performant controllers.
}

\newtext{
\textbf{\ourmethod uses physical knowledge to construct DR ranges.} \ourmethod takes advantage of the LLM's physical reasoning capabilities, which serve as a strong prior on DR ranges that are reasonable and intuitive. In the Appendix, we provide an example of the LLM output along with its explanations. We see that the LLM chooses the lower half of the RAPP restitution range, explaining that "restitution affects how the robot bounces off surfaces... lower range as we're not focusing on bouncing." For gravity, it chooses a relatively small range for "small tweaks to represent minor slopes or variations the robot might need to adapt to." Thus, \ourmethod proposes more reasonable DR configurations than CEM and BayRn, which treat DR as a numerical black-box optimization problem and relies on noisy real-world feedback for improving DR parameter proposal.
}

\begin{figure}
\centering
\includegraphics[width=0.45\columnwidth]{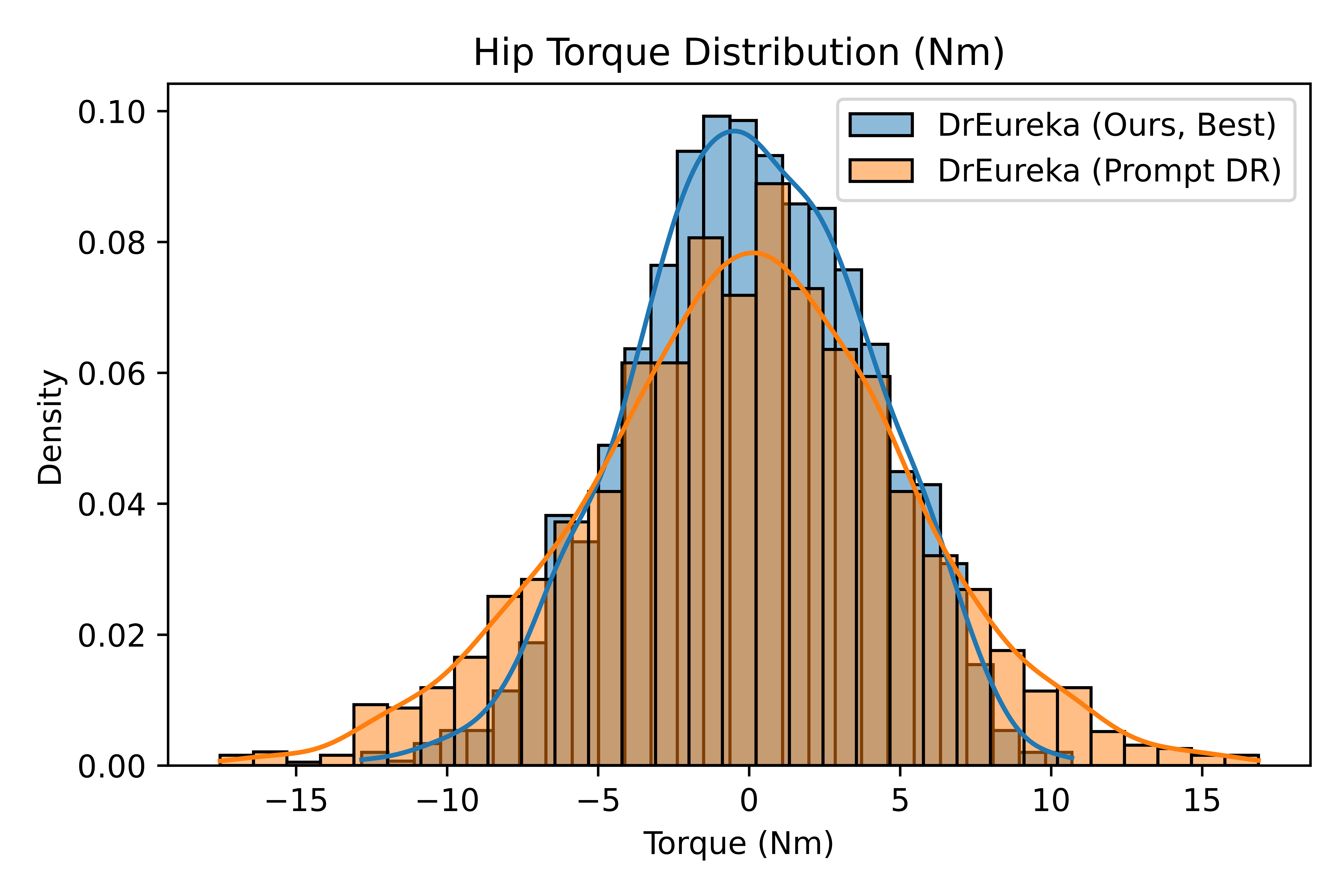}
\caption{Policies trained on \ourmethod DR configurations exert less torque in the real world.}
\label{figure:torque}
\end{figure}

\textbf{\ourmethod outperforms all DR ablations.} The real-world evaluation of these ablations is included in Table~\ref{table:main-results}. We first analyze the group of ablations that fix a single choice of DR configuration or lack thereof. We see that our tasks clearly demand domain randomization as \textbf{No DR} is inferior to both \ourmethod and \texttt{Human-Designed}. However, finding a suitable DR is not trivial. \textbf{Prompt DR} suggests wide parameter ranges (especially over friction as seen in Figure~\ref{figure:dr-prompt-example}) that forces the robot to over-exert forces; this result is validated in Figure~\ref{figure:torque} where we visualize the histogram of hip torque readings from real-world deployment of \ourmethod policies versus \textbf{Prompt DR} policies. On the other hand, using \textbf{Human-Designed DR} does not match the performance of \ourmethod, illustrating the importance of reward-aware domain randomization. Onto the sampling-based baselines, the subpar performance of \textbf{Random Sampling} suggests the effectiveness of LLMs as hypothesis generators, consistent with prior works that have found LLMs to be effective for suggesting initial samples for optimization problems~\citep{yang2023large,zhang2023using,anonymous2024large,ma2023eureka,romera2023mathematical}. However, fully utilizing LLM's zero-shot generation capability requires proper grounding of the sampling space. \textbf{No Prior} and \textbf{Uninformative Prior}, despite using a LLM as sampler, performs very poorly and often results in policies that trigger safety protection power cutoff in the real world. One common concern for LLM-based solutions is data leakage, in which the LLM has seen the problems and solutions for an evaluation task. In our setting, if the LLM has seen the simulations tasks and consequently the \texttt{human-designed} ranges in the open-sourced code base, then even if the priors are withheld in the context, it should be possible to output reasonable ranges out of the box. Fortunately, the negative results of \textbf{No Prior} \newtext{confirm} that data leakage does not appear in our evaluation. Altogether, these results affirm that both reward-aware parameter priors and LLM as a hypothesis generator in the \ourmethod framework are necessary for \newtext{the} best real-world performance.

\subsection{The Walking Globe Trick}

\jdtext{Our experiments above have focused on a thorough validation of \ourmethod on \newtext{existing tasks}, where a state-of-the-art sim-to-real approach was readily available as a reference point. Having validated \ourmethod's ability to automate sim-to-real to comparable performance levels with human design, we now employ \ourmethod for a challenging new task. In circus performances, the walking globe trick involves a performer balancing on a large sphere. Inspired by this, we train our quadruped to walk on an inflated yoga ball. The deformable and bouncy nature of the yoga ball complicates this task as IsaacGym simulation does not permit faithfully modeling the resulting complex dynamics of quadruped motion on the ball.} See Figure~\ref{figure:yoga} for visualizations of the simulation and the real world environment. This is a novel and arguably more difficult task than most solved quadrupedal tasks. Naturally, there is no pre-existing sim-to-real reward function or domain randomization configuration, making this task an ideal test-bed for \ourmethod's ability to accelerate robot skill discovery.

\begin{table}
\small
\centering
\resizebox{\columnwidth}{!}{
  \begin{tabular}{l|c}
  \toprule
  Walking Globe & Time on ball (s) \\
  \midrule
  Simulation & 10.7 $\pm$ 5.20 \\
  Real (Lab setting with center point support) & 15.4 $\pm$ 4.17 \\
  \bottomrule
 \end{tabular}}
  \caption{\textbf{\ourmethod results on walking globe.} In both simulation and the real world, the \ourmethod policy can balance and walk on the yoga ball for longer than 10 seconds.}
  \label{table:globe-walking}
\end{table}

The simulation environment is adapted from~\citet{ji2023dribblebot}, which simulates the Go1 playing with a small soccer ball. The robot and the ball are allowed to move around a large plane in simulation, but in the \newtext{lab setting} for safety, we limit the robot's movement by strapping it to a center support point in the room that is kept at the same height as the robot's center of mass; we use a human as the center point (see left of Figure~\ref{figure:yoga}) in order to provide support in the case of falling. The ball is free to move within a radius of 1 meter around this point. We perform \ourmethod using the same hyperparameters as in the locomotion task and report policy performance in Table~\ref{table:globe-walking}; we include the \ourmethod reward function and DR configurations for this task in Appendix. We observe the quadruped staying on the ball for an average of 15.43 seconds in the real world, many times making recovery actions to stabilize the ball and readjust its pose. In simulation, since the policy experiences a wide range of randomization parameters and perturbations, its average episode length is 10.72 seconds.

\newtext{Furthermore, given that the lab environment has limited floor space, we also deploy our policy on diverse, uncontrolled outdoor real-world scenes to further test the policy's robustness. With appropriate controls that limits the speed of the robot, the policy operated effectively for over four minutes under various conditions. Notably, the robot demonstrated stable navigation on grass, adeptly handled transitions over height obstacles, and moved smoothly onto sidewalks and wooden bridges. We also tested the policy's robustness by introducing perturbations such as kicking the yoga ball and operating the policy as the ball deflated. In all scenarios, the policy successfully managed these challenges, showcasing its adaptability and robustness across diverse operational conditions. See project website for videos.}

 In summary, \ourmethod's adeptness at tackling the novel and complex task of quadrupedal globe walking showcases its capacity to push the boundaries of what is achievable in robotic control tasks. This feat, achieved without prior specific sim-to-real pipelines, highlights \ourmethod's potential as a \newtext{versatile} tool in accelerating the development and deployment of robust robotic policies in the real world.

\section{Conclusion} 
\label{sec:conclusion}

We have presented \ourmethod, a novel technique for using large language models to guide sim-to-real reinforcement learning. Without human supervision, \ourmethod can automatically generate effective reward functions and domain randomization configurations comparable to human-crafted ones. \ourmethod is validated on \newtext{quadrupedal locomotion and dexterous manipulation}, and we have shown its potential in solving novel challenging tasks such as quadruped globe walking. We believe that \ourmethod demonstrates the potential of accelerating robot learning research by using foundation models to automate the difficult design aspects of low-level skill learning. 

\section{Limitations}
\label{sec:limitations}
While \ourmethod demonstrates the potential of leveraging LLMs for automating the sim-to-real transfer process in robotics, there are several areas of improvement to the current implementation:
\begin{itemize}
  \item \textbf{Lack of visual inputs}: The current implementation of \ourmethod does not integrate visual data from the environment, relying instead on lower-dimensional state representations. Incorporating vision-based inputs could potentially improve the robustness and generalizability of the learned policies in the real world, where visual cues play a critical role in navigation and interaction.

  \item \textbf{Static domain randomization parameters}: In the current framework, the domain randomization (DR) parameters, once generated, remain fixed during policy training. Dynamic adjustment of DR parameters based on policy performance or real-world feedback could further improve the sim-to-real transferability.
  \item \textbf{Lack of policy selection mechanism}: The evaluation of \ourmethod primarily focuses on the effectiveness of the generated reward functions and DR configurations. However, a systematic approach for selecting the most promising policies out of the generated candidates for real-world deployment is not explored. Integrating a mechanism that predicts real-world efficacy based on simulation performance or other heuristics could streamline the process of identifying the best policies for deployment.
\end{itemize}

\section*{\arxiv{Acknowledgements}}
We thank Gabe Margolis and Ge Yang for their assistance on our quadrupedal platform, Ankur Handa and Viktor Makoviychuk for their assistance on NVIDIA Isaac Gym simulator. We acknowledge funding support from NSF CAREER Award 2239301, ONR award N00014-22-1-2677, NSF Award CCF-1917852, and ARO Award W911NF-20-1-0080.
\clearpage

\bibliographystyle{unsrtnat}
\bibliography{references}

\begin{thebibliography}{70}
\providecommand{\natexlab}[1]{#1}
\providecommand{\url}[1]{\texttt{#1}}
\expandafter\ifx\csname urlstyle\endcsname\relax
  \providecommand{\doi}[1]{doi: #1}\else
  \providecommand{\doi}{doi: \begingroup \urlstyle{rm}\Url}\fi

\bibitem[Ahn et~al.(2022)Ahn, Brohan, Brown, Chebotar, Cortes, David, Finn, Fu, Gopalakrishnan, Hausman, et~al.]{ahn2022can}
Michael Ahn, Anthony Brohan, Noah Brown, Yevgen Chebotar, Omar Cortes, Byron David, Chelsea Finn, Chuyuan Fu, Keerthana Gopalakrishnan, Karol Hausman, et~al.
\newblock Do as i can, not as i say: Grounding language in robotic affordances.
\newblock \emph{arXiv preprint arXiv:2204.01691}, 2022.

\bibitem[Singh et~al.(2023)Singh, Blukis, Mousavian, Goyal, Xu, Tremblay, Fox, Thomason, and Garg]{singh2023progprompt}
Ishika Singh, Valts Blukis, Arsalan Mousavian, Ankit Goyal, Danfei Xu, Jonathan Tremblay, Dieter Fox, Jesse Thomason, and Animesh Garg.
\newblock Progprompt: Generating situated robot task plans using large language models.
\newblock In \emph{2023 IEEE International Conference on Robotics and Automation (ICRA)}, pages 11523--11530. IEEE, 2023.

\bibitem[Huang et~al.(2023{\natexlab{a}})Huang, Xia, Shah, Driess, Zeng, Lu, Florence, Mordatch, Levine, Hausman, et~al.]{huang2023grounded}
Wenlong Huang, Fei Xia, Dhruv Shah, Danny Driess, Andy Zeng, Yao Lu, Pete Florence, Igor Mordatch, Sergey Levine, Karol Hausman, et~al.
\newblock Grounded decoding: Guiding text generation with grounded models for robot control.
\newblock \emph{arXiv preprint arXiv:2303.00855}, 2023{\natexlab{a}}.

\bibitem[Liang et~al.(2023)Liang, Huang, Xia, Xu, Hausman, Ichter, Florence, and Zeng]{liang2023code}
Jacky Liang, Wenlong Huang, Fei Xia, Peng Xu, Karol Hausman, Brian Ichter, Pete Florence, and Andy Zeng.
\newblock Code as policies: Language model programs for embodied control.
\newblock In \emph{2023 IEEE International Conference on Robotics and Automation (ICRA)}, pages 9493--9500. IEEE, 2023.

\bibitem[Brohan et~al.(2023)Brohan, Brown, Carbajal, Chebotar, Chen, Choromanski, Ding, Driess, Dubey, Finn, et~al.]{brohan2023rt}
Anthony Brohan, Noah Brown, Justice Carbajal, Yevgen Chebotar, Xi~Chen, Krzysztof Choromanski, Tianli Ding, Danny Driess, Avinava Dubey, Chelsea Finn, et~al.
\newblock Rt-2: Vision-language-action models transfer web knowledge to robotic control.
\newblock \emph{arXiv preprint arXiv:2307.15818}, 2023.

\bibitem[Kwon et~al.(2023)Kwon, Di~Palo, and Johns]{kwon2023language}
Teyun Kwon, Norman Di~Palo, and Edward Johns.
\newblock Language models as zero-shot trajectory generators.
\newblock \emph{arXiv preprint arXiv:2310.11604}, 2023.

\bibitem[Yu et~al.(2023)Yu, Gileadi, Fu, Kirmani, Lee, Arenas, Chiang, Erez, Hasenclever, Humplik, et~al.]{yu2023language}
Wenhao Yu, Nimrod Gileadi, Chuyuan Fu, Sean Kirmani, Kuang-Huei Lee, Montse~Gonzalez Arenas, Hao-Tien~Lewis Chiang, Tom Erez, Leonard Hasenclever, Jan Humplik, et~al.
\newblock Language to rewards for robotic skill synthesis.
\newblock \emph{arXiv preprint arXiv:2306.08647}, 2023.

\bibitem[Xie et~al.(2023{\natexlab{a}})Xie, Zhao, Wu, Liu, Luo, Zhong, Yang, and Yu]{xie2023text2reward}
Tianbao Xie, Siheng Zhao, Chen~Henry Wu, Yitao Liu, Qian Luo, Victor Zhong, Yanchao Yang, and Tao Yu.
\newblock Text2reward: Automated dense reward function generation for reinforcement learning.
\newblock \emph{arXiv preprint arXiv:2309.11489}, 2023{\natexlab{a}}.

\bibitem[Ma et~al.(2023)Ma, Liang, Wang, Huang, Bastani, Jayaraman, Zhu, Fan, and Anandkumar]{ma2023eureka}
Yecheng~Jason Ma, William Liang, Guanzhi Wang, De-An Huang, Osbert Bastani, Dinesh Jayaraman, Yuke Zhu, Linxi Fan, and Anima Anandkumar.
\newblock Eureka: Human-level reward design via coding large language models.
\newblock \emph{arXiv preprint arXiv:2310.12931}, 2023.

\bibitem[Andrychowicz et~al.(2020)Andrychowicz, Baker, Chociej, Jozefowicz, McGrew, Pachocki, Petron, Plappert, Powell, Ray, et~al.]{andrychowicz2020learning}
OpenAI:~Marcin Andrychowicz, Bowen Baker, Maciek Chociej, Rafal Jozefowicz, Bob McGrew, Jakub Pachocki, Arthur Petron, Matthias Plappert, Glenn Powell, Alex Ray, et~al.
\newblock Learning dexterous in-hand manipulation.
\newblock \emph{The International Journal of Robotics Research}, 39\penalty0 (1):\penalty0 3--20, 2020.

\bibitem[{\AA}str{\"o}m and Eykhoff(1971)]{aastrom1971system}
Karl~Johan {\AA}str{\"o}m and Peter Eykhoff.
\newblock System identification—a survey.
\newblock \emph{Automatica}, 7\penalty0 (2):\penalty0 123--162, 1971.

\bibitem[Jaquier et~al.(2023)Jaquier, Welle, Gams, Yao, Fichera, Billard, Ude, Asfour, and Kragi{\'c}]{jaquier2023transfer}
No{\'e}mie Jaquier, Michael~C Welle, Andrej Gams, Kunpeng Yao, Bernardo Fichera, Aude Billard, Ale{\v{s}} Ude, Tamim Asfour, and Danica Kragi{\'c}.
\newblock Transfer learning in robotics: An upcoming breakthrough? a review of promises and challenges.
\newblock \emph{arXiv preprint arXiv:2311.18044}, 2023.

\bibitem[Muratore et~al.(2022)Muratore, Ramos, Turk, Yu, Gienger, and Peters]{muratore2022robot}
Fabio Muratore, Fabio Ramos, Greg Turk, Wenhao Yu, Michael Gienger, and Jan Peters.
\newblock Robot learning from randomized simulations: A review.
\newblock \emph{Frontiers in Robotics and AI}, page~31, 2022.

\bibitem[Tobin et~al.(2017)Tobin, Fong, Ray, Schneider, Zaremba, and Abbeel]{tobin2017domain}
Josh Tobin, Rachel Fong, Alex Ray, Jonas Schneider, Wojciech Zaremba, and Pieter Abbeel.
\newblock Domain randomization for transferring deep neural networks from simulation to the real world.
\newblock In \emph{2017 IEEE/RSJ international conference on intelligent robots and systems (IROS)}, pages 23--30. IEEE, 2017.

\bibitem[Peng et~al.(2018)Peng, Andrychowicz, Zaremba, and Abbeel]{peng2018sim}
Xue~Bin Peng, Marcin Andrychowicz, Wojciech Zaremba, and Pieter Abbeel.
\newblock Sim-to-real transfer of robotic control with dynamics randomization.
\newblock In \emph{2018 IEEE international conference on robotics and automation (ICRA)}, pages 3803--3810. IEEE, 2018.

\bibitem[Vuong et~al.(2019)Vuong, Vikram, Su, Gao, and Christensen]{vuong2019pick}
Quan Vuong, Sharad Vikram, Hao Su, Sicun Gao, and Henrik~I Christensen.
\newblock How to pick the domain randomization parameters for sim-to-real transfer of reinforcement learning policies?
\newblock \emph{arXiv preprint arXiv:1903.11774}, 2019.

\bibitem[Kumar et~al.(2021)Kumar, Fu, Pathak, and Malik]{kumar2021rma}
Ashish Kumar, Zipeng Fu, Deepak Pathak, and Jitendra Malik.
\newblock Rma: Rapid motor adaptation for legged robots.
\newblock \emph{arXiv preprint arXiv:2107.04034}, 2021.

\bibitem[Wang et~al.(2023{\natexlab{a}})Wang, Duan, Fox, and Srinivasa]{wang2023newton}
Yi~Ru Wang, Jiafei Duan, Dieter Fox, and Siddhartha Srinivasa.
\newblock Newton: Are large language models capable of physical reasoning?
\newblock \emph{arXiv preprint arXiv:2310.07018}, 2023{\natexlab{a}}.

\bibitem[Yang et~al.(2023)Yang, Wang, Lu, Liu, Le, Zhou, and Chen]{yang2023large}
Chengrun Yang, Xuezhi Wang, Yifeng Lu, Hanxiao Liu, Quoc~V Le, Denny Zhou, and Xinyun Chen.
\newblock Large language models as optimizers.
\newblock \emph{arXiv preprint arXiv:2309.03409}, 2023.

\bibitem[Zhang et~al.(2023{\natexlab{a}})Zhang, Desai, Bae, Lorraine, and Ba]{zhang2023using}
Michael~R Zhang, Nishkrit Desai, Juhan Bae, Jonathan Lorraine, and Jimmy Ba.
\newblock Using large language models for hyperparameter optimization.
\newblock \emph{arXiv e-prints}, pages arXiv--2312, 2023{\natexlab{a}}.

\bibitem[Anonymous(2024)]{anonymous2024large}
Anonymous.
\newblock Large language models to enhance bayesian optimization, 2024.
\newblock URL \url{https://openreview.net/forum?id=OOxotBmGol}.

\bibitem[Romera-Paredes et~al.(2023)Romera-Paredes, Barekatain, Novikov, Balog, Kumar, Dupont, Ruiz, Ellenberg, Wang, Fawzi, et~al.]{romera2023mathematical}
Bernardino Romera-Paredes, Mohammadamin Barekatain, Alexander Novikov, Matej Balog, M~Pawan Kumar, Emilien Dupont, Francisco~JR Ruiz, Jordan~S Ellenberg, Pengming Wang, Omar Fawzi, et~al.
\newblock Mathematical discoveries from program search with large language models.
\newblock \emph{Nature}, pages 1--3, 2023.

\bibitem[Rudin et~al.(2022)Rudin, Hoeller, Reist, and Hutter]{rudin2022learning}
Nikita Rudin, David Hoeller, Philipp Reist, and Marco Hutter.
\newblock Learning to walk in minutes using massively parallel deep reinforcement learning.
\newblock In \emph{Conference on Robot Learning}, pages 91--100. PMLR, 2022.

\bibitem[Lee et~al.(2020)Lee, Hwangbo, Wellhausen, Koltun, and Hutter]{lee2020learning}
Joonho Lee, Jemin Hwangbo, Lorenz Wellhausen, Vladlen Koltun, and Marco Hutter.
\newblock Learning quadrupedal locomotion over challenging terrain.
\newblock \emph{Science robotics}, 5\penalty0 (47):\penalty0 eabc5986, 2020.

\bibitem[Margolis et~al.(2022)Margolis, Yang, Paigwar, Chen, and Agrawal]{margolis2022rapid}
Gabriel~B Margolis, Ge~Yang, Kartik Paigwar, Tao Chen, and Pulkit Agrawal.
\newblock Rapid locomotion via reinforcement learning.
\newblock \emph{arXiv preprint arXiv:2205.02824}, 2022.

\bibitem[Margolis and Agrawal(2023)]{margolis2023walk}
Gabriel~B Margolis and Pulkit Agrawal.
\newblock Walk these ways: Tuning robot control for generalization with multiplicity of behavior.
\newblock In \emph{Conference on Robot Learning}, pages 22--31. PMLR, 2023.

\bibitem[Akkaya et~al.(2019)Akkaya, Andrychowicz, Chociej, Litwin, McGrew, Petron, Paino, Plappert, Powell, Ribas, et~al.]{akkaya2019solving}
Ilge Akkaya, Marcin Andrychowicz, Maciek Chociej, Mateusz Litwin, Bob McGrew, Arthur Petron, Alex Paino, Matthias Plappert, Glenn Powell, Raphael Ribas, et~al.
\newblock Solving rubik's cube with a robot hand.
\newblock \emph{arXiv preprint arXiv:1910.07113}, 2019.

\bibitem[Handa et~al.(2023)Handa, Allshire, Makoviychuk, Petrenko, Singh, Liu, Makoviichuk, Van~Wyk, Zhurkevich, Sundaralingam, et~al.]{handa2023dextreme}
Ankur Handa, Arthur Allshire, Viktor Makoviychuk, Aleksei Petrenko, Ritvik Singh, Jingzhou Liu, Denys Makoviichuk, Karl Van~Wyk, Alexander Zhurkevich, Balakumar Sundaralingam, et~al.
\newblock Dextreme: Transfer of agile in-hand manipulation from simulation to reality.
\newblock In \emph{2023 IEEE International Conference on Robotics and Automation (ICRA)}, pages 5977--5984. IEEE, 2023.

\bibitem[Qi et~al.(2023)Qi, Kumar, Calandra, Ma, and Malik]{qi2023hand}
Haozhi Qi, Ashish Kumar, Roberto Calandra, Yi~Ma, and Jitendra Malik.
\newblock In-hand object rotation via rapid motor adaptation.
\newblock In \emph{Conference on Robot Learning}, pages 1722--1732. PMLR, 2023.

\bibitem[Shaw et~al.(2023)Shaw, Agarwal, and Pathak]{shaw2023leap}
Kenneth Shaw, Ananye Agarwal, and Deepak Pathak.
\newblock Leap hand: Low-cost, efficient, and anthropomorphic hand for robot learning.
\newblock \emph{arXiv preprint arXiv:2309.06440}, 2023.

\bibitem[Zhang et~al.(2023{\natexlab{b}})Zhang, Zhang, Pertsch, Liu, Ren, Chang, Sun, and Lim]{zhang2023bootstrap}
Jesse Zhang, Jiahui Zhang, Karl Pertsch, Ziyi Liu, Xiang Ren, Minsuk Chang, Shao-Hua Sun, and Joseph~J Lim.
\newblock Bootstrap your own skills: Learning to solve new tasks with large language model guidance.
\newblock \emph{arXiv preprint arXiv:2310.10021}, 2023{\natexlab{b}}.

\bibitem[Szot et~al.(2023)Szot, Schwarzer, Agrawal, Mazoure, Talbott, Metcalf, Mackraz, Hjelm, and Toshev]{szot2023large}
Andrew Szot, Max Schwarzer, Harsh Agrawal, Bogdan Mazoure, Walter Talbott, Katherine Metcalf, Natalie Mackraz, Devon Hjelm, and Alexander Toshev.
\newblock Large language models as generalizable policies for embodied tasks.
\newblock \emph{arXiv preprint arXiv:2310.17722}, 2023.

\bibitem[Tang et~al.(2023)Tang, Yu, Tan, Zen, Faust, and Harada]{tang2023saytap}
Yujin Tang, Wenhao Yu, Jie Tan, Heiga Zen, Aleksandra Faust, and Tatsuya Harada.
\newblock Saytap: Language to quadrupedal locomotion.
\newblock \emph{arXiv preprint arXiv:2306.07580}, 2023.

\bibitem[Wang et~al.(2023{\natexlab{b}})Wang, Gonzalez-Pumariega, Sharma, and Choudhury]{wang2023demo2code}
Huaxiaoyue Wang, Gonzalo Gonzalez-Pumariega, Yash Sharma, and Sanjiban Choudhury.
\newblock Demo2code: From summarizing demonstrations to synthesizing code via extended chain-of-thought.
\newblock \emph{arXiv preprint arXiv:2305.16744}, 2023{\natexlab{b}}.

\bibitem[Huang et~al.(2023{\natexlab{b}})Huang, Jiang, Dong, Qiao, Gao, and Li]{huang2023instruct2act}
Siyuan Huang, Zhengkai Jiang, Hao Dong, Yu~Qiao, Peng Gao, and Hongsheng Li.
\newblock Instruct2act: Mapping multi-modality instructions to robotic actions with large language model.
\newblock \emph{arXiv preprint arXiv:2305.11176}, 2023{\natexlab{b}}.

\bibitem[Wang et~al.(2023{\natexlab{c}})Wang, Xie, Jiang, Mandlekar, Xiao, Zhu, Fan, and Anandkumar]{wang2023voyager}
Guanzhi Wang, Yuqi Xie, Yunfan Jiang, Ajay Mandlekar, Chaowei Xiao, Yuke Zhu, Linxi Fan, and Anima Anandkumar.
\newblock Voyager: An open-ended embodied agent with large language models.
\newblock \emph{arXiv preprint arXiv:2305.16291}, 2023{\natexlab{c}}.

\bibitem[Liu et~al.(2023)Liu, Jiang, Zhang, Liu, Zhang, Biswas, and Stone]{liu2023llm+}
Bo~Liu, Yuqian Jiang, Xiaohan Zhang, Qiang Liu, Shiqi Zhang, Joydeep Biswas, and Peter Stone.
\newblock Llm+ p: Empowering large language models with optimal planning proficiency.
\newblock \emph{arXiv preprint arXiv:2304.11477}, 2023.

\bibitem[Silver et~al.(2023)Silver, Dan, Srinivas, Tenenbaum, Kaelbling, and Katz]{silver2023generalized}
Tom Silver, Soham Dan, Kavitha Srinivas, Joshua~B Tenenbaum, Leslie~Pack Kaelbling, and Michael Katz.
\newblock Generalized planning in pddl domains with pretrained large language models.
\newblock \emph{arXiv preprint arXiv:2305.11014}, 2023.

\bibitem[Ding et~al.(2023)Ding, Zhang, Paxton, and Zhang]{ding2023task}
Yan Ding, Xiaohan Zhang, Chris Paxton, and Shiqi Zhang.
\newblock Task and motion planning with large language models for object rearrangement.
\newblock \emph{arXiv preprint arXiv:2303.06247}, 2023.

\bibitem[Lin et~al.(2023)Lin, Agia, Migimatsu, Pavone, and Bohg]{lin2023text2motion}
Kevin Lin, Christopher Agia, Toki Migimatsu, Marco Pavone, and Jeannette Bohg.
\newblock Text2motion: From natural language instructions to feasible plans.
\newblock \emph{arXiv preprint arXiv:2303.12153}, 2023.

\bibitem[Xie et~al.(2023{\natexlab{b}})Xie, Yu, Zhu, Bai, Gong, and Soh]{xie2023translating}
Yaqi Xie, Chen Yu, Tongyao Zhu, Jinbin Bai, Ze~Gong, and Harold Soh.
\newblock Translating natural language to planning goals with large-language models.
\newblock \emph{arXiv preprint arXiv:2302.05128}, 2023{\natexlab{b}}.

\bibitem[Wang et~al.(2023{\natexlab{d}})Wang, Ling, Yuan, Shridhar, Bao, Qin, Wang, Xu, and Wang]{wang2023gensim}
Lirui Wang, Yiyang Ling, Zhecheng Yuan, Mohit Shridhar, Chen Bao, Yuzhe Qin, Bailin Wang, Huazhe Xu, and Xiaolong Wang.
\newblock Gensim: Generating robotic simulation tasks via large language models.
\newblock \emph{arXiv preprint arXiv:2310.01361}, 2023{\natexlab{d}}.

\bibitem[Wang et~al.(2023{\natexlab{e}})Wang, Xian, Chen, Wang, Wang, Fragkiadaki, Erickson, Held, and Gan]{wang2023robogen}
Yufei Wang, Zhou Xian, Feng Chen, Tsun-Hsuan Wang, Yian Wang, Katerina Fragkiadaki, Zackory Erickson, David Held, and Chuang Gan.
\newblock Robogen: Towards unleashing infinite data for automated robot learning via generative simulation.
\newblock \emph{arXiv preprint arXiv:2311.01455}, 2023{\natexlab{e}}.

\bibitem[Tiboni et~al.(2023)Tiboni, Klink, Peters, Tommasi, D'Eramo, and Chalvatzaki]{tiboni2023domain}
Gabriele Tiboni, Pascal Klink, Jan Peters, Tatiana Tommasi, Carlo D'Eramo, and Georgia Chalvatzaki.
\newblock Domain randomization via entropy maximization.
\newblock \emph{arXiv preprint arXiv:2311.01885}, 2023.

\bibitem[Mehta et~al.(2020)Mehta, Diaz, Golemo, Pal, and Paull]{mehta2020active}
Bhairav Mehta, Manfred Diaz, Florian Golemo, Christopher~J Pal, and Liam Paull.
\newblock Active domain randomization.
\newblock In \emph{Conference on Robot Learning}, pages 1162--1176. PMLR, 2020.

\bibitem[Ramos et~al.(2019)Ramos, Possas, and Fox]{ramos2019bayessim}
Fabio Ramos, Rafael~Carvalhaes Possas, and Dieter Fox.
\newblock Bayessim: adaptive domain randomization via probabilistic inference for robotics simulators.
\newblock \emph{arXiv preprint arXiv:1906.01728}, 2019.

\bibitem[Chebotar et~al.(2019)Chebotar, Handa, Makoviychuk, Macklin, Issac, Ratliff, and Fox]{chebotar2019closing}
Yevgen Chebotar, Ankur Handa, Viktor Makoviychuk, Miles Macklin, Jan Issac, Nathan Ratliff, and Dieter Fox.
\newblock Closing the sim-to-real loop: Adapting simulation randomization with real world experience.
\newblock In \emph{2019 International Conference on Robotics and Automation (ICRA)}, pages 8973--8979. IEEE, 2019.

\bibitem[Muratore et~al.(2021)Muratore, Eilers, Gienger, and Peters]{muratore2021data}
Fabio Muratore, Christian Eilers, Michael Gienger, and Jan Peters.
\newblock Data-efficient domain randomization with bayesian optimization.
\newblock \emph{IEEE Robotics and Automation Letters}, 6\penalty0 (2):\penalty0 911--918, 2021.

\bibitem[Xie et~al.(2021)Xie, Da, Van~de Panne, Babich, and Garg]{xie2021dynamics}
Zhaoming Xie, Xingye Da, Michiel Van~de Panne, Buck Babich, and Animesh Garg.
\newblock Dynamics randomization revisited: A case study for quadrupedal locomotion.
\newblock In \emph{2021 IEEE International Conference on Robotics and Automation (ICRA)}, pages 4955--4961. IEEE, 2021.

\bibitem[Yu et~al.(2017)Yu, Tan, Liu, and Turk]{yu2017preparing}
Wenhao Yu, Jie Tan, C~Karen Liu, and Greg Turk.
\newblock Preparing for the unknown: Learning a universal policy with online system identification.
\newblock \emph{arXiv preprint arXiv:1702.02453}, 2017.

\bibitem[Tan et~al.(2018)Tan, Zhang, Coumans, Iscen, Bai, Hafner, Bohez, and Vanhoucke]{tan2018sim}
Jie Tan, Tingnan Zhang, Erwin Coumans, Atil Iscen, Yunfei Bai, Danijar Hafner, Steven Bohez, and Vincent Vanhoucke.
\newblock Sim-to-real: Learning agile locomotion for quadruped robots.
\newblock \emph{arXiv preprint arXiv:1804.10332}, 2018.

\bibitem[Pinto et~al.(2017)Pinto, Davidson, Sukthankar, and Gupta]{pinto2017robust}
Lerrel Pinto, James Davidson, Rahul Sukthankar, and Abhinav Gupta.
\newblock Robust adversarial reinforcement learning.
\newblock In \emph{International Conference on Machine Learning}, pages 2817--2826. PMLR, 2017.

\bibitem[Nagabandi et~al.(2018)Nagabandi, Clavera, Liu, Fearing, Abbeel, Levine, and Finn]{nagabandi2018learning}
Anusha Nagabandi, Ignasi Clavera, Simin Liu, Ronald~S Fearing, Pieter Abbeel, Sergey Levine, and Chelsea Finn.
\newblock Learning to adapt in dynamic, real-world environments through meta-reinforcement learning.
\newblock \emph{arXiv preprint arXiv:1803.11347}, 2018.

\bibitem[Bousmalis et~al.(2018)Bousmalis, Irpan, Wohlhart, Bai, Kelcey, Kalakrishnan, Downs, Ibarz, Pastor, Konolige, et~al.]{bousmalis2018using}
Konstantinos Bousmalis, Alex Irpan, Paul Wohlhart, Yunfei Bai, Matthew Kelcey, Mrinal Kalakrishnan, Laura Downs, Julian Ibarz, Peter Pastor, Kurt Konolige, et~al.
\newblock Using simulation and domain adaptation to improve efficiency of deep robotic grasping.
\newblock In \emph{2018 IEEE international conference on robotics and automation (ICRA)}, pages 4243--4250. IEEE, 2018.

\bibitem[James et~al.(2019)James, Wohlhart, Kalakrishnan, Kalashnikov, Irpan, Ibarz, Levine, Hadsell, and Bousmalis]{james2019sim}
Stephen James, Paul Wohlhart, Mrinal Kalakrishnan, Dmitry Kalashnikov, Alex Irpan, Julian Ibarz, Sergey Levine, Raia Hadsell, and Konstantinos Bousmalis.
\newblock Sim-to-real via sim-to-sim: Data-efficient robotic grasping via randomized-to-canonical adaptation networks.
\newblock In \emph{Proceedings of the IEEE/CVF Conference on Computer Vision and Pattern Recognition}, pages 12627--12637, 2019.

\bibitem[Ren et~al.(2023)Ren, Dai, Burchfiel, and Majumdar]{ren2023adaptsim}
Allen~Z Ren, Hongkai Dai, Benjamin Burchfiel, and Anirudha Majumdar.
\newblock Adaptsim: Task-driven simulation adaptation for sim-to-real transfer.
\newblock \emph{arXiv preprint arXiv:2302.04903}, 2023.

\bibitem[Ha et~al.(2023)Ha, Florence, and Song]{ha2023scaling}
Huy Ha, Pete Florence, and Shuran Song.
\newblock Scaling up and distilling down: Language-guided robot skill acquisition.
\newblock \emph{arXiv preprint arXiv:2307.14535}, 2023.

\bibitem[Russell and Norvig(1995)]{russell1995artificial}
Stuart~J Russell and Peter Norvig.
\newblock \emph{Artificial Intelligence: A Modern Approach}.
\newblock Prentice Hall, Englewood Cliffs, NJ, USA, 1st edition, 1995.

\bibitem[Sutton and Barto(2018)]{sutton2018reinforcement}
Richard~S Sutton and Andrew~G Barto.
\newblock \emph{Reinforcement Learning: An Introduction}.
\newblock MIT Press, Cambridge, MA, USA, 2nd edition, 2018.

\bibitem[Booth et~al.(2023)Booth, Knox, Shah, Niekum, Stone, and Allievi]{booth2023perils}
Serena Booth, W~Bradley Knox, Julie Shah, Scott Niekum, Peter Stone, and Alessandro Allievi.
\newblock The perils of trial-and-error reward design: misdesign through overfitting and invalid task specifications.
\newblock In \emph{Proceedings of the AAAI Conference on Artificial Intelligence}, volume~37, pages 5920--5929, 2023.

\bibitem[Kim et~al.(2023)Kim, Oh, Lee, Choi, Ji, Jung, Youm, and Hwangbo]{kim2023not}
Yunho Kim, Hyunsik Oh, Jeonghyun Lee, Jinhyeok Choi, Gwanghyeon Ji, Moonkyu Jung, Donghoon Youm, and Jemin Hwangbo.
\newblock Not only rewards but also constraints: Applications on legged robot locomotion.
\newblock \emph{arXiv preprint arXiv:2308.12517}, 2023.

\bibitem[Ouyang et~al.(2022)Ouyang, Wu, Jiang, Almeida, Wainwright, Mishkin, Zhang, Agarwal, Slama, Ray, et~al.]{ouyang2022training}
Long Ouyang, Jeffrey Wu, Xu~Jiang, Diogo Almeida, Carroll Wainwright, Pamela Mishkin, Chong Zhang, Sandhini Agarwal, Katarina Slama, Alex Ray, et~al.
\newblock Training language models to follow instructions with human feedback.
\newblock \emph{Advances in Neural Information Processing Systems}, 35:\penalty0 27730--27744, 2022.

\bibitem[Chung et~al.(2014)Chung, Gulcehre, Cho, and Bengio]{chung2014empirical}
Junyoung Chung, Caglar Gulcehre, KyungHyun Cho, and Yoshua Bengio.
\newblock Empirical evaluation of gated recurrent neural networks on sequence modeling, 2014.

\bibitem[Makoviychuk et~al.(2021)Makoviychuk, Wawrzyniak, Guo, Lu, Storey, Macklin, Hoeller, Rudin, Allshire, Handa, et~al.]{makoviychuk2021isaac}
Viktor Makoviychuk, Lukasz Wawrzyniak, Yunrong Guo, Michelle Lu, Kier Storey, Miles Macklin, David Hoeller, Nikita Rudin, Arthur Allshire, Ankur Handa, et~al.
\newblock Isaac gym: High performance gpu-based physics simulation for robot learning.
\newblock \emph{arXiv preprint arXiv:2108.10470}, 2021.

\bibitem[OpenAI(2023)]{openai2023gpt4}
OpenAI.
\newblock Gpt-4 technical report, 2023.

\bibitem[Kroese et~al.(2013)Kroese, Rubinstein, and Glynn]{KROESE201319}
Dirk~P. Kroese, Reuven~Y. Rubinstein, and Peter~W. Glynn.
\newblock Chapter 2 - the cross-entropy method for estimation.
\newblock In C.R. Rao and Venu Govindaraju, editors, \emph{Handbook of Statistics}, volume~31 of \emph{Handbook of Statistics}, pages 19--34. Elsevier, 2013.
\newblock \doi{https://doi.org/10.1016/B978-0-444-53859-8.00002-3}.
\newblock URL \url{https://www.sciencedirect.com/science/article/pii/B9780444538598000023}.

\bibitem[De~Boer et~al.(2005)De~Boer, Kroese, Mannor, and Rubinstein]{CrossEntropy}
Pieter-Tjerk De~Boer, Dirk~P Kroese, Shie Mannor, and Reuven~Y Rubinstein.
\newblock A tutorial on the cross-entropy method.
\newblock \emph{Annals of operations research}, 134\penalty0 (1):\penalty0 19--67, 2005.

\bibitem[Frazier(2018)]{frazier2018tutorial}
Peter~I. Frazier.
\newblock A tutorial on bayesian optimization, 2018.

\bibitem[Schulman et~al.(2017)Schulman, Wolski, Dhariwal, Radford, and Klimov]{schulman2017proximal}
John Schulman, Filip Wolski, Prafulla Dhariwal, Alec Radford, and Oleg Klimov.
\newblock Proximal policy optimization algorithms.
\newblock \emph{arXiv preprint arXiv:1707.06347}, 2017.

\bibitem[Ji et~al.(2023)Ji, Margolis, and Agrawal]{ji2023dribblebot}
Yandong Ji, Gabriel~B Margolis, and Pulkit Agrawal.
\newblock Dribblebot: Dynamic legged manipulation in the wild.
\newblock \emph{arXiv preprint arXiv:2304.01159}, 2023.

\end{thebibliography}

\clearpage 

\onecolumn
\renewcommand\lstlistingname{Prompt}
\setcounter{lstlisting}{0}
\section*{Appendix}
\subsection*{A. Full Prompts and Success Criteria}
\label{appendix:full-prompts}
In this section, we provide \ourmethod prompts used for experiments and ablations. We also include the success criteria used to select reward candidates.
\subsection*{A1. Reward Generation Prompts}
This section contains the system and task prompts for generating reward functions for forward locomotion and globe walking tasks using \ourmethod.
\begin{lstlisting}[basicstyle=\fontfamily{\ttdefault}\scriptsize, breaklines=true,caption={\ourmethod system prompt for reward generation.}]
You are a reward engineer trying to write reward functions to solve reinforcement learning tasks as effective as possible.
Your goal is to write a reward function for the environment that will help the agent learn the task described in text. 
Your reward function should use useful variables from the environment as inputs. As an example,
the reward function signature can be: {task_reward_signature_string}
Make sure any new tensor or variable you introduce is on the same device as the input tensors. 
\end{lstlisting}

\begin{lstlisting}[basicstyle=\fontfamily{\ttdefault}\scriptsize, breaklines=true,caption={\ourmethod forward locomotion task prompt for reward generation. \arxiv{The safety instructions encouraging a steady torso and smooth movement result in natural gaits.}}]
To make the go1 quadruped run forward with a velocity of exactly 2.0 m/s in the positive x direction of the global coordinate frame. The policy will be trained in simulation and deployed in the real world, so the policy should be as steady and stable as possible with minimal action rate. Specifically, as it's running, the torso should remain near a z position of 0.34, and the orientation should be perpendicular to gravity. Also, the legs should move smoothly and avoid the DOF limits.
\end{lstlisting}

\begin{lstlisting}[basicstyle=\fontfamily{\ttdefault}\scriptsize, breaklines=true,caption={\newtext{\ourmethod cube rotation task prompt for reward generation. Note that we limit the encouraged rotation speed to 0.25 rad/s due to simulation inaccuracies causing extreme angular velocity measurements; while the policy is not explicitly rewarded for rotating faster than this, we find that faster rotation is implicitly rewarded by greater stability and consistency.}}]
To make the robot hand rotate the cube in hand with positive angular velocity around the z-axis as many times as possible. We want the cube to remain in the palm with near-zero linear velocity, and it should not fall out of the hand. The policy will be trained in simulation and deployed in the real world, so the policy should be smooth and steady, and the fingers should be penalized for deviating far from their initial position. For safety, we would like the cube to rotate at around 0.25 radians per second; however, it's okay if it rotates faster.
\end{lstlisting}

\begin{lstlisting}[basicstyle=\fontfamily{\ttdefault}\scriptsize, breaklines=true,caption={\ourmethod globe walking task prompt for reward generation. \arxiv{In our safety instruction, we provide a numerical value to ground torque scaling because \texttt{env.torques} in the code is the output of a black-box model rather than mathematically calculated.}}]
To make the go1 quadruped balance on the top of the ball. The quadruped should maintain a z-position of 2 * ball_radius or higher. Please keep in mind that the policy learned using your reward terms will be deployed on a robot in the real world. As such, you should prioritize safety, robustness, and feasibility over performance. Please generate reward terms that penalize actions that are unsafe or infeasible. Please also penalize jittery or fast actions that may burn out the motors. Also, remember to keep the scaling of your regularization terms small. If you choose to use env.torques, please keep in mind that this value will be large, so your scaling for this term should be near 0.00001.
\end{lstlisting}

\subsection*{A2. Reward Generation Ablation Prompts}
This section contains prompts used in ablation studies, specifically for generating reward functions without safety instructions to assess the impact of such instructions on the generated rewards.
\begin{lstlisting}[basicstyle=\fontfamily{\ttdefault}\scriptsize, breaklines=true,caption={\ourmethod forward locomotion task prompt for reward generation, without safety instructions.}]
The Python environment is {environment source code}. Write a reward function for the following task: To make the go1 quadruped run forward with a velocity of exactly 2.0 m/s in the positive x direction of the global coordinate frame.
\end{lstlisting}

\subsection*{A3. Domain Randomization Generation Prompts}
This section includes the initial system and user prompts for generating domain randomization configurations, demonstrating how \ourmethod is applied to different tasks for robust policy training.
\begin{lstlisting}[basicstyle=\fontfamily{\ttdefault}\scriptsize, breaklines=true,caption={\ourmethod system prompt for DR generation.}]
You are a reinforcement learning engineer. Your goal is to design a set of domain randomization parameters for the given task to facilitate successful deployment of the trained policy in the real world.
To do so, you will be given valid parameters as well as a range for each parameter that indicates the maximum and minimum values that parameter can take. Please note that your randomization ranges do not need to cover most of the range.
Also, you should keep in mind that the more you randomize, the more difficult it will be for the policy to learn the task within our fixed compute budget. A good policy should be trained only on randomization ranges that will help it adapt to the real world.
You should first reason over each parameter and determine if it's useful for domain randomization.
Then, you should output a range of values for each parameter that you think will be useful for the task in a real-world deployment. Please explain your reasoning for each parameter.

Output your response in the form of Python code that sets the parameters as variables, e.g.:
```
friction_range = [0.0, 1.0]
```
Please make your variable names match the parameter names provided. Each variable should be assigned a range formatted as a Python list with two elements. Write everything else as Python comments.
\end{lstlisting}

\begin{lstlisting}[basicstyle=\fontfamily{\ttdefault}\scriptsize, breaklines=true,caption={\ourmethod quadruped prompt with RAPP from \ourmethod policy. This prompt corresponds to the 'Our Method' configuration in Table \ref{table:main-results}.}]
The task is to train a quadruped robot to run on a variety of terrains indoor and outdoor. The goal of the robot is to run forward at 2.0 m/s while remaining steady and safe in the real world.
The robot will be trained in simulation and then deployed in the real world.
Our parameters and valid ranges are the following:
    friction_range = [0.0, 10.0]
    restitution_range = [0.0, 1.0]
    added_mass_range = [-5.0, 5.0]
    com_displacement_range = [-0.1, 0.1]
    motor_strength_range = [0.5, 2.0]
    Kp_factor_range = [0.5, 2.0]
    Kd_factor_range = [0.5, 2.0]
    dof_stiffness_range = [0.0, 1.0]
    dof_damping_range = [0.0, 0.5]
    dof_friction_range = [0.0, 0.01]
    dof_armature_range = [0.0, 0.01] (This is the range of values added onto the diagonal of the joint inertia matrix.)
    push_vel_xy_range = [0.0, 1.0]   (This is the range of magnitudes of a vector added onto the robot's xy velocity.)
    gravity_range = [-1.0, 1.0]      (This is the range of values added onto each dimension of [0.0, 0.0, -9.8].
                                      For example, [0.0, 0.0] would keep gravity constant.)
\end{lstlisting}

\begin{lstlisting}[basicstyle=\fontfamily{\ttdefault}\scriptsize, breaklines=true,caption={\newtext{\ourmethod cube rotation prompt with RAPP from \ourmethod policy. This prompt corresponds to the 'Our Method' configuration in Table \ref{table:leaphand-results}.}}]
The task is to train a robot hand to rotate the cube in hand. The goal of the robot is to rotate the cube with positive angular velocity around the z-axis as many times as possible while remaining steady and safe in the real world.
The robot will be trained in simulation and then deployed in the real world.
Our parameters and valid ranges are the following:
    HandFriction = [0.0, 10.0]
    HandRestitution = [0.0, 1.0]
    HandDOFStiffness = [1.0, 10.0]
    HandDOFDamping = [0.0, 0.5]
    HandDOFFriction = [0.0, 0.1]
    HandDOFArmature = [0.0, 0.01]
    ObjectMass = [0.01, 1.0]            # For reference, we measured the real cube to be 0.046 kg
    ObjectCOM = [-0.01, 0.01]
    ObjectFriction = [0.0, 10.0]
    ObjectRestitution = [0.0, 1.0]
\end{lstlisting}

\begin{lstlisting}[basicstyle=\fontfamily{\ttdefault}\scriptsize, breaklines=true,caption={\ourmethod globe walking prompt with RAPP from \ourmethod policy.}]
The task is to train a quadruped robot to balance on a yoga ball for as long as possible.
The robot will be trained in simulation and then deployed in the real world. Please note that our simulation environment models the ball as a solid rigid object, so the robot will not be able to deform the ball in any way. However, our real yoga ball is hollow, bouncy, and deformable, so the robot will need to adapt to this difference. Please keep this in mind when designing your domain randomization.
Our parameters and valid ranges are the following:
    robot_friction_range = [0.1, 1.0]
    robot_restitution_range = [0.0, 1.0]
    robot_payload_mass_range = [-1.0, 5.0]
    robot_com_displacement_range = [-0.1, 0.1]
    robot_motor_strength_range = [0.9, 1.1]
    robot_motor_offset_range = [-0.01, 0.1]
    ball_mass_range = [0.5, 5.0]
    ball_friction_range = [0.1, 3.0]
    ball_restitution_range = [0.0, 1.0]
    ball_drag_range = [0.0, 1.0]
    terrain_ground_friction_range = [0.0, 1.0]
    terrain_ground_restitution_range = [0.0, 1.0]
    terrain_tile_roughness_range = [0.0, 0.1]
    robot_push_vel_range = [0.0, 0.5]
    ball_push_vel_range = [0.0, 0.5]
    gravity_range = [-0.5, 0.5]
\end{lstlisting}

\subsection*{A4. Domain Randomization Generation Ablation Prompts}
This section includes prompts used in ablation experiments that test the importance of RAPP priors in the LLM prompt. Below, we include a prompt with no prior context and a prompt whose context is the entire range tested by the RAPP algorithm.

\begin{lstlisting}[basicstyle=\fontfamily{\ttdefault}\scriptsize, breaklines=true,caption={Initial quadruped prompt (no context). This prompt corresponds to the 'Without Prior' configuration in Table \ref{table:main-results}.}]
The task is to train a quadruped robot to run on a variety of terrains indoor and outdoor. The goal of the robot is to run forward at 2.0 m/s while remaining steady and safe in the real world.
The robot will be trained in simulation and then deployed in the real world.
Our parameters are the following:
    friction_range
    restitution_range
    added_mass_range
    com_displacement_range
    motor_strength_range
    Kp_factor_range
    Kd_factor_range
    dof_stiffness_range
    dof_damping_range
    dof_friction_range
    dof_armature_range        (This is the range of values added onto the diagonal of the joint inertia matrix.)
    push_vel_xy_range         (This is the range of magnitudes of a vector added onto the robot's xy velocity.)
    gravity_range             (This is the range of values added onto each dimension of [0.0, 0.0, -9.8]. For example,   
                              [0.0, 0.0] would keep gravity constant.)
\end{lstlisting}

\begin{lstlisting}[basicstyle=\fontfamily{\ttdefault}\scriptsize, breaklines=true,caption={Initial quadruped prompt (uninformative context). This prompt corresponds to the 'With Uninformative Prior' configuration in Table \ref{table:main-results}.}]
The task is to train a quadruped robot to run on a variety of terrains indoor and outdoor. The goal of the robot is to run forward at 2.0 m/s while remaining steady and safe in the real world.
The robot will be trained in simulation and then deployed in the real world.
Our parameters and valid ranges are the following:
    friction_range = [0.0, 10.0]
    restitution_range = [0.0, 1.0]
    added_mass_range = [-10.0, 10.0]
    com_displacement_range = [-10.0, 10.0]
    motor_strength_range = [0.0, 2.0]
    Kp_factor_range = [0.0, 2.0]
    Kd_factor_range = [0.0, 2.0]
    dof_stiffness_range = [0.0, 10.0]
    dof_damping_range = [0.0, 10.0]
    dof_friction_range = [0.0, 10.0]
    dof_armature_range = [0.0, 10.0]          (This is the range of values added onto the diagonal of the joint inertia matrix.)
    push_vel_xy_range = [0.0, 10.0]            (This is the range of magnitudes of a vector added onto the robot's xy velocity.)
    gravity_range = [-10.0, 10.0]               (This is the range of values added onto each dimension of [0.0, 0.0, -9.8]. For example, [0.0, 0.0] would keep gravity constant.)
\end{lstlisting}

\subsection*{A5. Success Criteria}
\begin{table}[H]
\centering
\resizebox{0.5\columnwidth}{!}{
  \begin{tabular}{l|cc}
  \toprule
  {Term} & Success Criteria \\
  \midrule  %
    Forward Locomotion & $\exp(-(v_x - v^t_x)^2 / 0.25)$ \\
    Cube Rotation & $\text{clip}(\omega_z, -0.25, 0.25)$ \\
    Walking Globe & $1$ \\
  \bottomrule
 \end{tabular}}
  \caption{\textbf{Success criteria for our tasks.} $v_x$ and $v_x^t$ are forward velocity and target velocity, respectively. $\omega_z$ is the angular velocity of the cube. The success criteria is summed over all steps within an episode.}
  \label{table:success-criteria}
\end{table}

\subsection*{B. \ourmethod Outputs}
\label{appendix:outputs}
\lstset{style=pythonstyle}
In this section, we detail the reward functions generated by \ourmethod and applied in the training of forward locomotion and globe walking task.

\subsection*{B1. LLM-Generated Rewards}

\begin{lstlisting}[basicstyle=\fontfamily{\ttdefault}\scriptsize, breaklines=true,caption={Final reward for forward locomotion task from \ourmethod.}]
def compute_reward(self):
    env = self.env  # Do not skip this line. Afterwards, use env.{parameter_name} to access parameters of the environment.

    # Ideal forward velocity in the x direction
    target_velocity_x = 2.0
    # Ideal height of the robot's torso
    target_height_z = 0.34

    # Compute the velocity reward component
    current_velocity_x = env.root_states[:, 7]  # Linear velocity in x from the root_states tensor
    velocity_error = torch.abs(current_velocity_x - target_velocity_x)
    velocity_reward = torch.exp(-velocity_error)

    # Compute the height reward component
    current_height = env.root_states[:, 2]  # Position in z from the root_states tensor
    height_error = torch.abs(current_height - target_height_z)
    height_reward = torch.exp(-5.0 * height_error)  # More weight to maintain height

    # Compute the orientation reward component
    # Ideal orientation is perpendicular to gravity, i.e., the projected gravity vector should be [0, 0, -1] in the robot's frame
    ideal_projected_gravity = torch.tensor([0., 0., -1.], device=env.device).repeat((env.num_envs, 1))
    orientation_error = torch.norm(env.projected_gravity - ideal_projected_gravity, dim=1)
    orientation_reward = torch.exp(-5.0 * orientation_error)  # More weight to maintain orientation

    # Legs movement within DOF limits reward component
    dof_limit_violations = torch.any(
        (env.dof_pos < env.dof_pos_limits[:, 0]) | (env.dof_pos > env.dof_pos_limits[:, 1]),
        dim=-1)
    dof_limit_violations_reward = 1.0 - dof_limit_violations.float()  # Penalize if any DOF limit is violated

    # Smoothness reward component (penalize the change in actions to encourage smooth movements)
    action_difference = torch.norm(env.actions - env.last_actions, dim=1)
    smoothness_reward = torch.exp(-0.1 * action_difference)

    # Combine reward components
    total_reward = velocity_reward * height_reward * orientation_reward * dof_limit_violations_reward * smoothness_reward

    # Debug information
    reward_components = {"velocity_reward": velocity_reward,
                         "height_reward": height_reward,
                         "orientation_reward": orientation_reward,
                         "dof_limit_violations_reward": dof_limit_violations_reward,
                         "smoothness_reward": smoothness_reward}
    return total_reward, reward_components
\end{lstlisting}

\begin{lstlisting}[basicstyle=\fontfamily{\ttdefault}\scriptsize, breaklines=true,caption={\newtext{Final reward for cube rotation task from \ourmethod.}}]
@torch.jit.script
def compute_reward(object_pos: torch.Tensor, object_angvel_finite_diff: torch.Tensor, object_linvel: torch.Tensor, leap_hand_dof_pos: torch.Tensor, init_pose_buf: torch.Tensor, reset_z_threshold: float) -> Tuple[torch.Tensor, Dict[str, torch.Tensor]]:
    # Constants for tuning the reward
    z_axis_index = 2
    ang_vel_target = 0.25
    max_ang_vel_reward = 2.5
    linear_velocity_penalty_coefficient = -3.0
    angular_velocity_penalty_coefficient = -2.5
    fall_penalty = -5.0
    deviation_penalty_coefficient = -0.2

    # Reward for positive angular velocity around the z-axis.
    ang_z_vel = object_angvel_finite_diff[:, z_axis_index]
    ang_z_vel_reward = torch.where(ang_z_vel > ang_vel_target, 
                                   ang_vel_target + (1 - torch.exp(ang_vel_target - ang_z_vel)), 
                                   ang_z_vel)

    ang_z_vel_reward = torch.clamp(ang_z_vel_reward, max=max_ang_vel_reward)

    # Penalize linear velocity to ensure the cube remains steady
    lin_vel_penalty = linear_velocity_penalty_coefficient * torch.norm(object_linvel, dim=1)

    # Penalize the cube falling out of the hand
    object_fall_penalty = torch.where(object_pos[:, z_axis_index] < reset_z_threshold,
                                      fall_penalty * torch.ones_like(object_pos[:, 0]),
                                      torch.zeros_like(object_pos[:, 0]))

    # Penalize deviation from initial finger positions
    deviation_penalty = deviation_penalty_coefficient * torch.norm(leap_hand_dof_pos - init_pose_buf, dim=1)

    # Total reward
    total_reward = ang_z_vel_reward + lin_vel_penalty + object_fall_penalty + deviation_penalty

    # Reward components dictionary for potential debugging or analysis
    reward_components = {
        "angular_velocity_reward": ang_z_vel_reward,
        "linear_velocity_penalty": lin_vel_penalty,
        "fall_penalty": object_fall_penalty,
        "deviation_penalty": deviation_penalty
    }

    return total_reward, reward_components
\end{lstlisting}

\begin{lstlisting}[basicstyle=\fontfamily{\ttdefault}\scriptsize, breaklines=true,caption={Final reward for globe walking task from \ourmethod. Due to a limitation in the original environment's codebase, the Eureka reward format here splits each term into a separate function and computes the final reward as a sum of all terms. Besides a minimal change in the prompt to describe this format, everything else is the same.}]
def _reward_height(self):
    env = self.env
    height_threshold = 2.0 * env.ball_radius
    height_temperature = 7.0  # Fine-tuned temperature parameter
    height_exp = torch.exp((env.base_pos[:, 2] - height_threshold) / height_temperature)
    height_reward = torch.where(env.base_pos[:, 2] >= height_threshold, height_exp, torch.zeros_like(env.base_pos[:, 2]))
    return 1.5 * height_reward  # Updated scaling

def _reward_balance(self):
    env = self.env
    balance_temperature = 5.0  # Fine-tuned temperature parameter
    ball_top = env.object_pos_world_frame.clone()
    ball_top[:, 2] += env.ball_radius

    feet_dist_to_ball_top = torch.norm(env.foot_positions - ball_top.unsqueeze(1), dim=-1)
    balance_exp = torch.exp(-feet_dist_to_ball_top / balance_temperature)
    balance_reward = torch.mean(balance_exp, dim=-1)
    return 2.0 * balance_reward  # Updated scaling

def _reward_smooth_actions(self):
    env = self.env
    action_diff = env.actions - env.last_actions
    smooth_actions_reward = -torch.mean(torch.abs(action_diff), dim=-1)
    return 1.0 * smooth_actions_reward  # Increase scale of smooth_actions_reward

def _reward_penalize_large_actions(self):
    env = self.env
    large_action_penalty = -torch.mean(torch.abs(env.actions), dim=-1)
    return 0.3 * large_action_penalty  # Increase scaling for penalize_large_actions
\end{lstlisting}

\subsection*{B2. LLM-Generated Domain Randomizations}
\renewcommand\lstlistingname{Generation}
\setcounter{lstlisting}{0}
In this section, we provide the examples of domain randomization configurations generated by \ourmethod given Reward-Aware Physics Prior.

\begin{lstlisting}[basicstyle=\fontfamily{\ttdefault}\scriptsize, breaklines=true,caption={Example \ourmethod DR configuration for forward locomotion.}]
# Friction is important as it affects how the robot interacts with different surfaces (indoor, outdoor).
friction_range = [0.5, 5.0]  # Moderate range to cover various surfaces like tiles, grass, dirt, etc.

# Restitution affects how the robot bounces off surfaces or objects; however, for a running task, this might be less critical.
restitution_range = [0.0, 0.5]  # Lower range as we're not focusing on bouncing, but it's still relevant for minor impacts.

# Added mass simulates the effect of carrying additional weight, which could influence stability and motor strength requirements.
added_mass_range = [-2.0, 2.0]  # A moderate range to simulate carrying light payloads or none.

# Center of mass displacement affects stability and maneuverability.
com_displacement_range = [-0.05, 0.05]  # Slight displacements to ensure the model can handle minor imbalances.

# Motor strength directly affects the robot's ability to move and respond to commands.
motor_strength_range = [0.75, 1.75]  # A moderate range allowing for variability in motor performance.

# Kp_factor (proportional gain) and Kd_factor (derivative gain) influence the control loop's responsiveness and stability.
Kp_factor_range = [0.75, 1.5]  # A range that allows for experimenting with control responsiveness without becoming unstable.
Kd_factor_range = [0.75, 1.5]  # Similar to Kp to ensure controllability under various conditions.

# Degrees of freedom (DOF) stiffness and damping affect how rigid and shock-absorbing the joints are, respectively.
dof_stiffness_range = [0.2, 0.8]  # A range that allows for some variation in joint stiffness without being too rigid or floppy.
dof_damping_range = [0.1, 0.4]  # Provides variability in how the robot absorbs shocks and returns to equilibrium.

# DOF friction and armature values are very specific parameters that could affect energy efficiency and precision in movements.
dof_friction_range = [0.0, 0.005]  # Small range based on the minor impact it has on performance but still considering efficiency.
dof_armature_range = [0.0, 0.005]  # Minimally varies the joint inertia to see the effects on stability and precision.

# Push velocity in the xy plane simulates external forces or disturbances, e.g., wind or collisions.
push_vel_xy_range = [0.0, 0.5]  # A range to help the robot learn to stabilize against moderate disturbances.

# Gravity variations can simulate different loading conditions or slight inclines/declines the robot might face.
gravity_range = [-0.2, 0.2]  # Small tweaks to represent minor slopes or variations the robot might need to adapt to.

# Note: Each range is chosen with the thought of exposing the robot to a variety of conditions it might face in the real world
# without making the task unnecessarily difficult to learn within our compute budget.
\end{lstlisting}

\begin{lstlisting}[basicstyle=\fontfamily{\ttdefault}\scriptsize, breaklines=true,caption={\newtext{Example \ourmethod DR configuration for cube rotation.}}]
# HandFriction determines how much grip the hand has on objects. Given real-world materials can vary,
# we should include a range, but it shouldn't be too broad to maintain gripping capability while ensuring learning feasibility.
randomizeHandFrictionLower: 0.5
randomizeHandFrictionUpper: 5.0

# HandRestitution reflects the bounce when the hand hits something, which is less critical for this specific task of rotating a cube,
# yet minor variations could simulate real-world scenarios where the hand might interact with the cube with varying intensities.
randomizeHandRestitutionLower: 0.0
randomizeHandRestitutionUpper: 0.2

# HandDOFStiffness affects how forcefully the joints can move, which is crucial for controlling the cube rotation.
# A moderate range should allow the policy to adapt to different stiffness levels found in real hardware.
randomizeHandDOFStiffnessLower: 4.0
randomizeHandDOFStiffnessUpper: 8.0

# HandDOFDamping influences how movements are smoothed or resisted by the joints, affecting the control precision.
# Including a range here helps ensure that the policy can deal with varying damping in real hands.
randomizeHandDOFDampingLower: 0.1
randomizeHandDOFDampingUpper: 0.4

# HandDOFFriction affects the internal joint movements. Since real robotic hands might have varying joint frictions,
# a small range can simulate these differences without compromising the hand's basic functionality.
randomizeHandDOFFrictionLower: 0.01
randomizeHandDOFFrictionUpper: 0.05

# HandDOFArmature refers to the inertia of the hand joints, which could slightly vary in reality but typically remains low.
# We'll randomize it minimally as it's less critical for cube rotation but still worth considering for physical accuracy.
randomizeHandDOFArmatureLower: 0.001
randomizeHandDOFArmatureUpper: 0.005

# ObjectMass is directly measured, but slight variations in mass distribution or accuracy of the measurement can occur.
# Thus, randomizing around the measured value simulates handling cubes of slightly different masses.
randomizeObjectMassLower: 0.04
randomizeObjectMassUpper: 0.05

# ObjectCOM (Center of Mass) might shift slightly in real objects depending on manufacturing variances.
# Small randomization here can help the policy cope with such variances in balancing the cube during rotation.
randomizeObjectCOMLower: -0.005
randomizeObjectCOMUpper: 0.005

# ObjectFriction affects how easily the cube rotates within the grasp. Since surface materials can vary greatly,
# a broader range here assists in preparing the policy for different cube surfaces.
randomizeObjectFrictionLower: 1.0
randomizeObjectFrictionUpper: 8.0

# ObjectRestitution represents how bouncy the cube is when dropped or thrown, which is less relevant for rotations,
# but slight variations might simulate interactions with the environment more realistically.
randomizeObjectRestitutionLower: 0.0
randomizeObjectRestitutionUpper: 0.3
\end{lstlisting}

\begin{lstlisting}[basicstyle=\fontfamily{\ttdefault}\scriptsize, breaklines=true,caption={Example \ourmethod DR configuration for globe walking.}]
# Friction between the robot and the ball is crucial because it affects how well the robot can balance and maneuver on the ball. Since the real ball can be less predictable, a wider range should promote adaptability.
robot_friction_range = [0.1, 1.0] 

# Restitution, or bounciness, will affect how the robot interacts with surfaces upon collision. Though the simulation does not account for ball deformation, varying restitution can simulate the unpredictability of these interactions.
robot_restitution_range = [0.2, 0.8] 

# Since the payload will directly affect the robot's balance and how it responds to shifts in weight, we allow for variability but avoid extreme negative values to maintain realism.
robot_payload_mass_range = [0.0, 3.0]

# Center of mass displacement affects balance and stability. Randomization within a moderate range can prepare the robot for shifts in its own weight distribution.
robot_com_displacement_range = [-0.05, 0.05]

# Motor strength is critical for moving and balancing. A narrow range ensures the robot remains capable of movement but can adapt to variability in its actuation power.
robot_motor_strength_range = [0.95, 1.05] 

# Motor offsets will simulate imperfections in actuator performance. Randomizing this could prepare the robot for real-world inaccuracies.
robot_motor_offset_range = [-0.005, 0.05]

# The ball's mass will significantly impact how the robot interacts with it. Since the ball is hollow and can be deformed, a middle-range should provide a good balance between too light and too heavy.
ball_mass_range = [1.0, 3.0]

# Ball friction and restitution are critical for preparing the robot to interact with a bouncy and deformable ball. These ranges allow for significant variability.
ball_friction_range = [0.5, 2.5]
ball_restitution_range = [0.4, 0.9] 

# Ball drag simulates air resistance, which could affect interactions at higher speeds.
ball_drag_range = [0.1, 0.5]

# The robot might not always operate on similar terrains, so simulating a range of frictions can be beneficial. However, the restitution of the ground is less critical here.
terrain_ground_friction_range = [0.2, 0.8]
terrain_ground_restitution_range = [0.0, 0.5] 

# Terrain roughness could influence balance and traction, so a slight variation can introduce realistic challenges without overwhelming the learning process.
terrain_tile_roughness_range = [0.02, 0.08]

# Varying the push velocities can help the robot learn to maintain balance against unexpected forces.
robot_push_vel_range = [0.1, 0.4]
ball_push_vel_range = [0.1, 0.4]

# Considering the task does not involve drastic changes in gravity, we only slightly vary this to simulate minor differences in weight sensation.
gravity_range = [-0.1, 0.1]
\end{lstlisting}
    
\begin{figure}[H]
\centering
\includegraphics[width=\columnwidth]{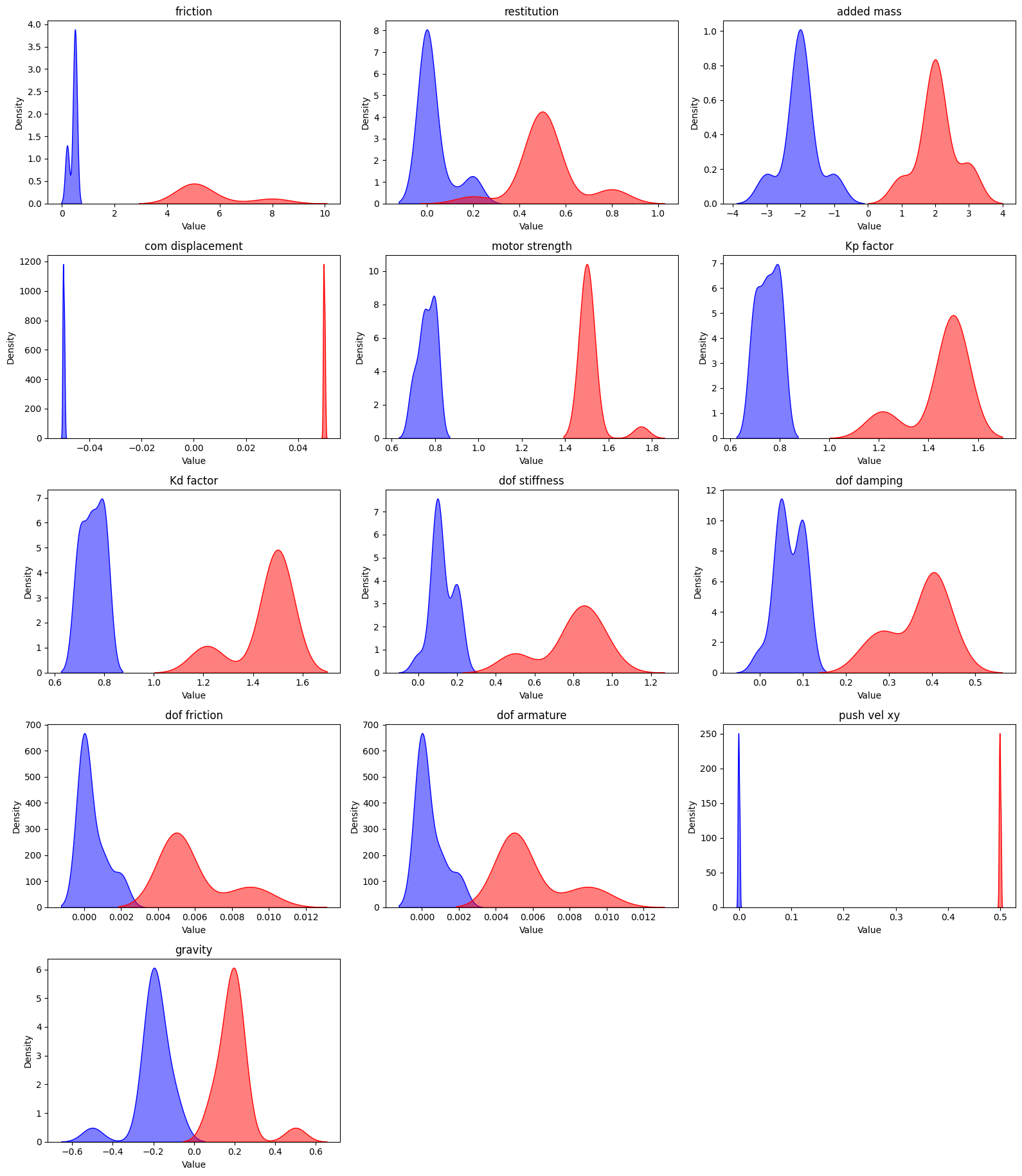}
\vspace{-0.4cm}
\caption{\newtext{Visualization of DR parameter ranges sampled by \ourmethod for forward locomotion: Blue represents the lower bound of the sampled DR parameter range and red represents the upper bound of the sampled DR parameter range. As shown, the LLM generates a series of diverse yet reasonable ranges. We also provide the training curves to further illustrate the difference between configurations.}}
\label{figure:dreureka-sampled-values}
\vspace{-0.2in}
\end{figure}

\begin{figure}[H]
\label{fig:unitree}\includegraphics[width=\columnwidth]{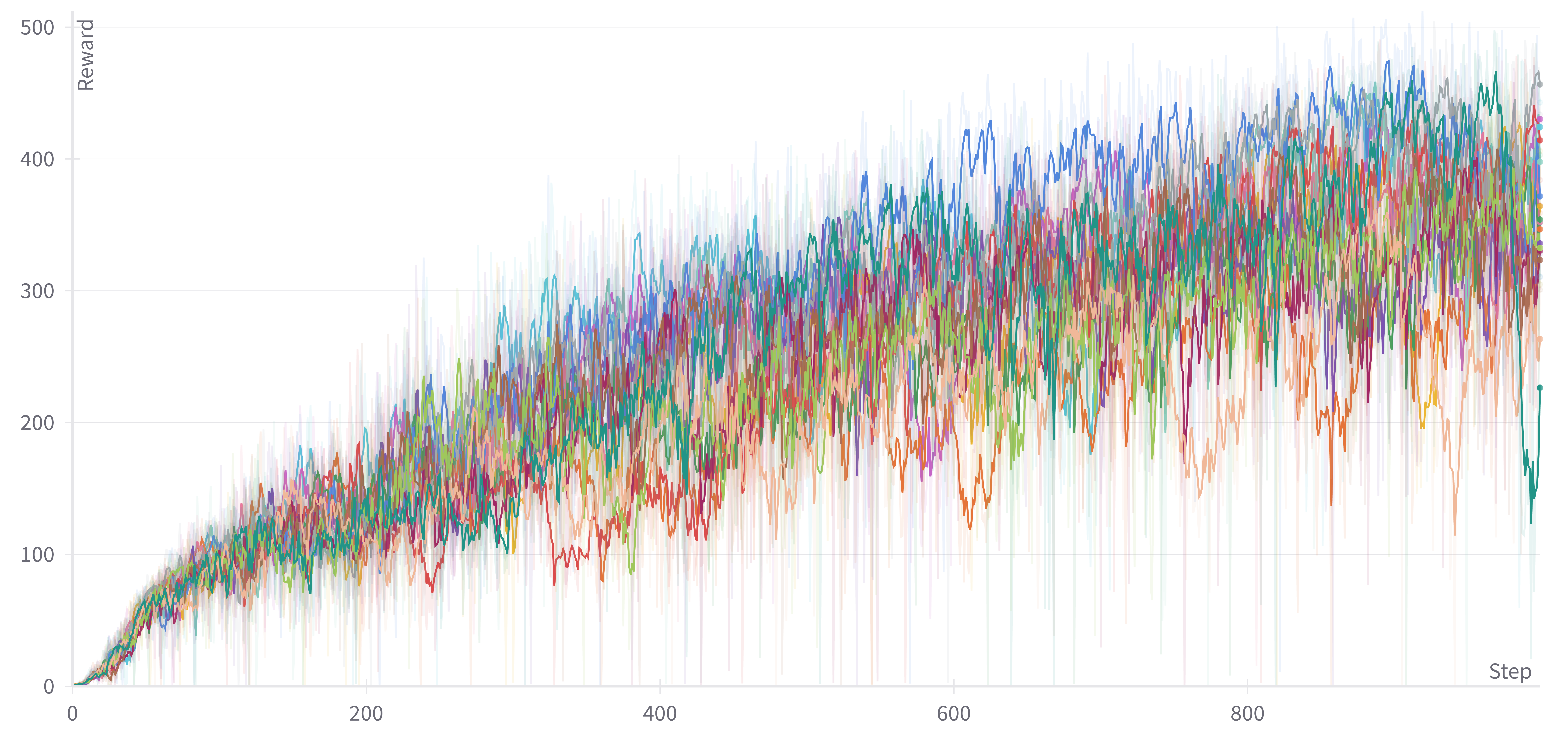}
\vspace{-0.4cm}
\caption{\newtext{Forward locomotion training curves for 16 DR configurations. All runs are trained with the same reward function.}}
\label{figure:dr-training-curves}
\vspace{-0.2in}
\end{figure}

\subsection*{\newtext{B3. LLM Reward Reflection}}
\newtext{The following is an example of reward reflection on the challenging Walking Globe task. Provided with task scores and other training statistics, the LLM is asked to evolve better reward functions.}

\begin{lstlisting}[basicstyle=\fontfamily{\ttdefault}\scriptsize, breaklines=true,caption={\newtext{Example \ourmethod reward reflection for globe walking. By modifying the scale for one term and changing the functional form of another, the LLM increases the task score average from 115 to 407.}}]
Eureka Output:
Here are the reward terms I am planning to implement:
1. height_reward: Encourage the agent to maintain a z-position of at least 2 * ball_radius.
2. balance_reward: Encourage the agent to balance on the ball by minimizing the distance between the feet and the top of the ball.
3. smooth_actions_reward: Encourage smooth actions to reduce jittery or fast actions.

```python
def _reward_height(self):
    env = self.env
    height_threshold = 2.0 * env.ball_radius
    height_reward = torch.where(env.base_pos[:, 2] >= height_threshold, torch.ones_like(env.base_pos[:, 2]), torch.zeros_like(env.base_pos[:, 2]))
    return 1.0 * height_reward

def _reward_balance(self):
    env = self.env
    ball_top = env.object_pos_world_frame + torch.tensor([0.0, 0.0, env.ball_radius], device=env.device).unsqueeze(0)
    feet_dist_to_ball_top = torch.norm(env.foot_positions - ball_top.unsqueeze(1), dim=-1)
    balance_reward = 1.0 - torch.mean(feet_dist_to_ball_top, dim=-1)
    return 1.0 * balance_reward

def _reward_smooth_actions(self):
    env = self.env
    action_diff = env.actions - env.last_actions
    smooth_actions_reward = -torch.mean(torch.abs(action_diff), dim=-1)
    return 0.1 * smooth_actions_reward
```

The height_reward motivates the agent to maintain a z-position above the threshold. The balance_reward ensures the agent learns to balance on the ball by minimizing the distance between the feet and the top of the ball. The smooth_actions_reward encourages the agent to take smoother actions rather than jittery or too fast actions that may burn out the motors.

####################################################################################################

Reward Reflection:
We trained a RL policy using the provided reward function code and tracked the values of the individual components in the reward function as well as global policy metrics such as task score and episode lengths after every 500 epochs and the maximum, mean, minimum values encountered:
task score: ['12.42', '62.72', '67.46', '69.36', '70.67', '80.36', '112.15', '152.94', '185.67', '241.25'], Max: 285.75, Mean: 115.54, Min: 12.42 
episode length: ['11.45', '62.74', '67.35', '69.35', '70.68', '79.95', '111.51', '152.29', '184.50', '237.52'], Max: 276.88, Mean: 115.86, Min: 11.45 
rew balance: ['8.28', '39.74', '42.03', '42.49', '43.45', '50.95', '75.72', '108.42', '135.23', '179.91'], Max: 217.18, Mean: 80.06, Min: 8.28 
rew height: ['12.29', '51.43', '55.16', '57.51', '58.48', '66.92', '95.47', '135.24', '167.38', '222.18'], Max: 267.01, Mean: 100.62, Min: 12.29 
rew smooth actions: ['-1.30', '-9.17', '-11.89', '-13.57', '-15.27', '-19.87', '-34.40', '-47.73', '-56.97', '-72.17'], Max: -1.30, Mean: -31.26, Min: -87.12 
rew total: ['19.26', '81.99', '85.29', '86.43', '86.66', '98.00', '136.80', '195.93', '245.64', '329.93'], Max: 397.07, Mean: 149.43, Min: 19.26 
Please carefully analyze the policy feedback and provide a new, improved reward function that can better solve the task. Some helpful tips for analyzing the policy feedback:
    (1) If the task score is always near zero, then you must rewrite the entire reward function
    (2) If the values for a certain reward component are near identical throughout, then this means RL is not able to optimize this component as it is written. You may consider
        (a) Changing its scale or the value of its temperature parameter
        (b) Re-writing the reward component 
        (c) Discarding the reward component
    (3) If some reward components magnitude is significantly larger, then you must re-scale its value to a proper range
Please analyze each existing reward component in the suggested manner above first, and then write the reward function code. The output of the reward function should be only your reward value.
The code output should be formatted as a python code string: "```python ... ```".

Some helpful tips for writing the reward function code:
    (1) You may find it helpful to normalize the reward to a fixed range by applying transformations like torch.exp to the overall reward or its components
    (2) If you choose to transform a reward component, then you must also introduce a temperature parameter inside the transformation function; this parameter must be a named variable in the reward function and it must not be an input variable. Each transformed reward component should have its own temperature variable
    (3) Most importantly, the reward codes must only use attributes of the provided environment object (namely, variables that have prefix env.). Under no circumstance can you introduce new input variables.

####################################################################################################

Eureka Output:
From the policy feedback, it seems that the reward for height ('rew height') and balance ('rew balance') are significant and show improvement, but the reward for smooth actions ('rew smooth actions') needs more optimization. I will increase the scale for smooth_actions_reward and introduce an exponential reward component for height_reward so that the reward gradient is smoother. Here is the updated reward function:

```python
def _reward_height(self):
    env = self.env
    height_threshold = 2.0 * env.ball_radius
    height_temperature = 10.0  # new temperature parameter for exponential transformation
    height_exp = torch.exp((env.base_pos[:, 2] - height_threshold) / height_temperature)
    height_reward = torch.where(env.base_pos[:, 2] >= height_threshold, height_exp, torch.zeros_like(env.base_pos[:, 2]))
    return 2.0 * height_reward  # increase scale of height_reward

def _reward_balance(self):
    env = self.env
    ball_top = env.object_pos_world_frame + torch.tensor([0.0, 0.0, env.ball_radius], device=env.device).unsqueeze(0)
    feet_dist_to_ball_top = torch.norm(env.foot_positions - ball_top.unsqueeze(1), dim=-1)
    balance_reward = 1.0 - torch.mean(feet_dist_to_ball_top, dim=-1)
    return 1.0 * balance_reward

def _reward_smooth_actions(self):
    env = self.env
    action_diff = env.actions - env.last_actions
    smooth_actions_reward = -torch.mean(torch.abs(action_diff), dim=-1)
    return 0.5 * smooth_actions_reward  # increase scale of smooth_actions_reward
```

In this updated reward function, I introduced a temperature parameter (`height_temperature`) and applied an exponential transformation to the height_reward so that the reward gradient is smoother. I also multiplied the height_reward scale by 2 to make it more significant. Finally, I increased the scale of the smooth_actions_reward by 0.5 to encourage smoother actions more than before, which will help further improve the policy.

####################################################################################################

Reward Reflection:
We trained a RL policy using the provided reward function code and tracked the values of the individual components in the reward function as well as global policy metrics such as task score and episode lengths after every 500 epochs and the maximum, mean, minimum values encountered:
task score: ['12.42', '65.13', '71.19', '76.92', '100.98', '320.91', '602.55', '741.82', '865.37', '924.64'], Max: 1040.59, Mean: 407.71, Min: 12.42 
episode length: ['11.45', '65.14', '71.13', '76.80', '100.93', '330.90', '608.89', '767.77', '903.78', '903.98'], Max: 1074.86, Mean: 409.03, Min: 11.45 
rew balance: ['8.28', '41.57', '44.23', '47.64', '64.50', '237.34', '464.35', '574.94', '673.07', '722.61'], Max: 811.20, Mean: 310.16, Min: 8.28 
rew height: ['25.01', '108.71', '122.09', '133.27', '181.14', '620.04', '1187.27', '1468.28', '1718.10', '1837.45'], Max: 2076.23, Mean: 797.53, Min: 25.01 
rew smooth actions: ['-6.52', '-35.68', '-38.78', '-42.31', '-66.38', '-271.25', '-508.06', '-624.86', '-740.29', '-808.17'], Max: -6.52, Mean: -340.10, Min: -911.98 
rew total: ['26.76', '114.60', '127.55', '138.60', '179.26', '586.13', '1143.56', '1418.36', '1650.88', '1751.89'], Max: 1975.46, Mean: 767.58, Min: 26.76 

...
\end{lstlisting}

\subsection*{C. Mathematical Representation of \ourmethod Rewards}
\label{appendix:math-rewards}
In this section, we convert the programmatic human-written and LLM-generated reward functions into mathematical expressions for comparison.

\begin{table}[H]
\centering
\resizebox{0.8\textwidth}{!}{
  \begin{tabular}{l|l}
  \toprule
  Symbol & Explanation \\
  \midrule
  \( v^t_x, v_x\) & Agent's and target's linear velocity along the x-axis. \\
  \( \omega^t_z, \omega_z\) & Agent's and target's angular velocity around the z-axis. \\
  \(v_z\) & Velocity along the z-axis. \\
  \(\omega_{xy}\) & Velocities in the roll and pitch directions. \\
  \(p^t_z, p_z\) & Agent's and target's base height. \\
  \(g_{xy}\) & Base orientation in the horizontal plane. \\
  \(j, j_l, j_h\) & Joint position and lower, upper joint limits. \\
  \(\tau\) & Applied torques. \\
  \(\ddot{j}\) & Joint acceleration. \\
  \(a_t, a_{t-1}\) & Consecutive actions to measure smoothness and action rate. \\
  \(t_{air}\) & Feet airtime during next contact transitions. \\
  \(foot\_position, ball\_top\_position\) & 3D Positions of the robot foot and the top of the ball. \\
  \bottomrule
  \end{tabular}
}
\caption{Explanation of symbols used in \newtext{forward locomotion} reward, Tables \ref{table:human-written-reward-function-forwardloco}, \ref{table:eureka-reward-function}, \ref{table:globe_walking_dr_parameters}.}
\label{table:symbol-explanations}
\end{table}

\begin{table}[H]
\centering
\begin{minipage}{0.45\linewidth}
\resizebox{\columnwidth}{!}{
  \begin{tabular}{l|cc}
  \toprule
  Reward Component & Math Expression \\
  \midrule
    \texttt{Linear velocity tracking} & $0.02 \cdot \exp\{-(v_x - v^t_x)^2 / 0.25 \}$ \\
    \texttt{Angular velocity tracking} & $0.01 \cdot \exp\{-(\omega_z - \omega^t_z)^2 / 0.25 \}$ \\
    \texttt{Z-velocity penalty} & $-0.04 \cdot v_z^2$ \\
    \texttt{Roll-pitch-velocity penalty} & $-0.001 \cdot \vert \omega_{xy} \vert^2$ \\
    \texttt{Base height penalty} & $-0.6 \cdot (p_z - p^t_z)^2$ \\
    \texttt{Base orientation penalty} & $-0.1 \cdot \vert g_{xy} \vert^2$ \\
    \texttt{Collision penalty} & $-0.02 \cdot \mathbf{1}[\text{collision}]$ \\
    \texttt{Joint limit penalty} & $-0.2 \cdot (\max(0, j_l - j) + \max(0, j - j_h))$ \\
    \texttt{Torque penalty} & $-2e-6 \cdot \vert \tau \vert^2$ \\
    \texttt{Joint acceleration penalty} & $-5e-9 \cdot \vert \ddot{j} \vert^2$ \\
    \texttt{Action rate penalty} & $-2e-4 \cdot \vert a_t - a_{t-1} \vert^2$ \\
    \texttt{Feet airtime} & $0.02 \cdot \sum t_{air} \cdot \mathbf{1}[\text{next contact}]$ \\
  \bottomrule
  \end{tabular}
}
\caption{\textbf{Human-written reward function for forward locomotion.} The total reward is the sum of the components above.}
\label{table:human-written-reward-function-forwardloco}
\end{minipage}
\hfill
\begin{minipage}{0.50\linewidth}
\resizebox{\columnwidth}{!}{
  \begin{tabular}{l|cc}
  \toprule
  Reward Component & Math Expression \\
  \midrule
    \texttt{Forward velocity} & $\exp\{ -(v_x - v^t_x)^2 / 2 \}$ \\
    \texttt{Action smoothness} & $-0.25 \cdot \vert a_t - a_{t-1} \vert$ \\
    \texttt{Angular velocity} & $-0.25 \cdot \Vert \omega_{xyz} \Vert_2$ \\
    \texttt{Eureka reward} & \texttt{Forward velocity} + \texttt{Action smoothness} \\ & + \texttt{Angular velocity} \\
  \bottomrule
  \end{tabular}
}
\caption{\textbf{Final reward for forward locomotion from Eureka without safety instruction.}}
\label{table:eureka-reward-function}
\end{minipage}
\end{table}

\begin{table}[H]
\centering
\resizebox{0.7\columnwidth}{!}{
  \begin{tabular}{l|cc}
  \toprule
  Reward Component & Math Expression  \\
  \midrule  %
    \texttt{Height} & $ 1.5 \cdot \mathbb{I}_{\{p^t_z > p_z\}} \cdot \exp\{\frac{p^t_z - p_z}{7}\}$ \\
    \texttt{Balance} & $2 \cdot \exp\{\frac{- \Vert foot\_position - ball\_top\_position \Vert}{5} \}$ \\
    \texttt{Action smoothness} & $-1 \cdot \vert a_t - a_{t-1} \vert$ \\
    \texttt{Large Action Penalty} & $-0.3 \cdot | a_t | $ \\
    \texttt{Eureka reward} & \texttt{Height} + \texttt{Balance} + \\
    & \texttt{Action smoothness} + \texttt{Large Action Penalty}\\
  \bottomrule
 \end{tabular}}
  \caption{\textbf{Final reward for the walking globe task.}}
  \label{table:yoga_ball_reward}
\end{table}

\begin{table}[H]
\centering
\begin{minipage}{0.45\textwidth}
\centering
\begin{tabular}{l|l}
\toprule
Symbol & Explanation \\
\midrule
\( p_z \) & Height of the object. \\
\( \omega_z \) & Angular velocity vector of the object along the z-axis. \\
\( \mathbf{v} \) & Linear velocity vector of the object. \\
\( \mathbf{q} \) & Current joint angles of the hand. \\
\( \mathbf{q}_0 \) & Initial joint angles of the hand. \\
\( z_{\text{threshold}} \) & Threshold z-axis position below which the object is considered fallen. \\
\( \alpha \) & Target angular velocity around the z-axis. Set to 0.25. \\
\(\tau_i\) & Applied torque of motor i. \\
\(W \) & Work done by the motors. \\
\bottomrule
\end{tabular}
\caption{Explanations of symbols used in cube rotation reward.}
\label{table:symbols}
\end{minipage}%
\hfill
\begin{minipage}{0.45\textwidth}
\centering
\begin{tabular}{cl}
\toprule
\textbf{Reward Component} & \textbf{Formula} \\
\midrule
Angular Velocity Reward & \( 1.25 \cdot \text{clip}( \omega_z, -0.25, 0.25)\) \\
Linear Velocity Penalty \(P_{v}\) & \( -0.3 \cdot \|\mathbf{v}\|_1 \) \\
Pose Difference Penalty \(P_d\) & \( -0.1 \|\mathbf{q} - \mathbf{q}_0\| \) \\
Torque Penalty \(P_{\text{torque}}\) & \( -0.1 \cdot \text{sum}(\tau_i^2 \)) \\
Work Penalty \(P_{\text{work}}\) & \(-1 \cdot W \) \\
Object Falling Penalty \(P_f\) & \(\begin{cases} -10 & \text{if } p_z < z_{\text{threshold}} \\ 0 & \text{otherwise} \end{cases} \) \\
\bottomrule
\end{tabular}
\caption{Human-written reward function for cube rotation.}
\label{table:cube_rot_human}
\end{minipage}
\end{table}

\begin{table}[H]
\centering
\begin{tabular}{l|c}
\toprule
    Reward Component & Math Expression \\
\midrule
Angular Velocity Reward $R_{\omega_z}$ & 
$\min \left( 
\begin{cases}
\alpha + (1 - \exp\{\alpha - \omega_z\}) & \text{if } \omega_z > \alpha \\
\omega_z & \text{otherwise}
\end{cases}, 2.5 \right)$ \\
Linear Velocity Penalty $P_v$ & $-3 \cdot \|\mathbf{v}\|$ \\
Object Falling Penalty $P_f$ & 
$\begin{cases} 
-5 & \text{if } p_z < z_{\text{threshold}} \\ 
0 & \text{otherwise} 
\end{cases}$ \\
Pose Difference Penalty $P_d$ & $-0.2 \|\mathbf{q} - \mathbf{q}_0\|$ \\
Eureka reward & $R_{\omega_z} + P_v + P_f + P_d$ \\
\bottomrule
\end{tabular}
\caption{\textbf{Final reward for cube rotation task from \textit{our method}.}}
\label{table:cube_rot_reward_dreureka}
\end{table}

\subsection*{D. Experimental Setup}
\label{appendix:experiment-setup}

\subsection*{D1. Forward Locomotion}
For the forward locomotion task, our policy takes joint positions, joint velocities, a gravity vector, and a history of past observations and actions as input. It produces joint position commands for a PD controller, which has a proportional gain of 20 and derivative gain of 0.5.

We extend the simulation setup from \citet{margolis2022rapid}, and we include additional domain randomization parameters, specifically joint stiffness, damping, friction, and armature that were not in the their work. These parameters, along with the others in Table~\ref{table:forward_loco_dr_parameters}, were randomized during training. We chose these parameters based on IsaacGym's documentation on rigid body, rigid shape, and DOF properties\footnotemark[2].

\begin{table}[H]
\centering
\resizebox{0.7\columnwidth}{!}{
  \begin{tabular}{l | c c}
  \toprule
  Property & Valid Range & RAPP Search Range \\
  \midrule
  \texttt{friction} & $[0, \infty)$ & $[0, 10]$ \\
  \texttt{restitution} & $[0, 1]$ & $[0, 1]$ \\
  \texttt{payload mass} & $(-\infty, \infty)$ & $[-10, 10]$ \\
  \texttt{center of mass displacement} & $(-\infty, \infty)$ & $[-10, 10]$ \\
  \texttt{motor strength} & $[0, \infty)$ & $[0, 2]$ \\
  \texttt{scaling factors for proportional gain} & $[0, \infty)$ & $[0, 2]$ \\
  \texttt{scaling factors for derivative gain} & $[0, \infty)$ & $[0, 2]$ \\
  \texttt{push velocity} & $[0, \infty)$ & $[0, 10]$ \\
  \texttt{gravity} & $(-\infty, \infty)$ & $[-10, 10]$ \\
  \midrule
  \texttt{dof stiffness} & $[0, \infty)$ & $[0, 10]$ \\
  \texttt{dof damping} & $[0, \infty)$ & $[0, 10]$ \\
  \texttt{dof friction} & $[0, \infty)$ & $[0, 10]$ \\
  \texttt{dof armature} & $[0, \infty)$ & $[0, 10]$ \\
  \bottomrule
 \end{tabular}}
  \caption{\textbf{Domain randomization parameters for forward locomotion, along with their valid ranges and RAPP search ranges.} Though the scale of these parameters differs, each RAPP range is chosen from one of four general-purpose ranges (\texttt{0\_to\_infty}, \texttt{0\_to\_1}, \texttt{centered\_0}, \texttt{centered\_1}).}
  \label{table:forward_loco_dr_parameters}
\end{table}

\footnotetext[2]{Relevant functions in the documentation are \texttt{isaacgym.gymapi.RigidBodyProperties}, \texttt{isaacgym.gymapi.RigidShapeProperties}, \texttt{isaacgym.gymapi.Gym.get\_actor\_dof\_properties()}. Note that among these properties, there are a few fields that we found had no effect in simulation. We discarded them for our domain randomization.}

\subsection*{\newtext{D2. Cube Rotation}}
\vspace{-0.5em}
\newtext{For the cube rotation task, we follow the training and deployment workflow outlined by the LeapHand authors. For training all the policies, we use the same GRU architecture that receives 16 joint angles as input and outputs 16 target joint angles. We also follow the LeapHand training code to randomize the initial pose of the hand and the size of the cube. When deploying trained policies in the real world, the target joint angles are passed as position commands to a PID controller running at 20 Hz.}

\newtext{In addition to the initial pose of the hand and the size of the cube, the \texttt{Human Designed} policy is trained with DR in object mass, object center of mass, hand friction, stiffness and damping. In \ourmethod, we extend the simulation setup to include additional domain randomization parameters, such as hand restitution, joint friction, armature, object friction and object restitution. These parameters, along with the others, are detailed in Table \ref{table:cube_rot_dr_parameters}.}

\begin{table}[H]
\centering
\resizebox{0.7\columnwidth}{!}{
  \begin{tabular}{l | c c}
  \toprule
  Property & Valid Range & RAPP Search Range \\
  \midrule
  \texttt{object mass} & $[0, \infty)$ & $[0.01, 1]$ \\
  \texttt{object center of mass} & $[0, \infty)$ & $[-0.01, 0.01]$ \\
  \texttt{hand friction} & $[0, \infty)$ & $[0, 10]$ \\
  \texttt{dof stiffness} & $[0, \infty)$ & $[1, 10]$ \\
  \texttt{dof damping} & $[0, \infty)$ & $[0, 0.5]$ \\
  \midrule
  \texttt{hand restitution} & $[0, 1]$ & $[0, 1]$ \\
  \texttt{dof friction} & $[0, \infty)$ & $[0, 0.1]$ \\
  \texttt{armature} & $[0, \infty)$ & $[0, 0.01]$ \\
  \texttt{object friction} & $[0, \infty)$ & $[0, 10]$ \\
  \texttt{object restitution} & $[0, 1]$ & $[0, 1]$ \\
  \bottomrule
 \end{tabular}}
  \caption{\newtext{\textbf{Domain randomization parameters for cube rotation, along with their valid ranges and RAPP search ranges.}}}
  \label{table:cube_rot_dr_parameters}
\end{table}

\subsection*{D3. Globe Walking}
For globe walking, we largely extend the framework from forward locomotion, with a few exceptions. First, the policy takes in an additional yaw sensor as input. Second, to account for actuator inaccuracies in the real world, we use an actuator network from \citet{ji2023dribblebot}; this network is pretrained on log data to predict real robot torques from joint commands, and we use it to compute torques from actions in simulation when training the quadruped. Third, we have additional domain randomization parameters, shown in Table~\ref{table:globe_walking_dr_parameters}.

In the real world, we deploy our quadruped on a 34-inch yoga ball. We did not have a stable pole to tether our quadruped, so we instead resort to a human holding the end of the leash; however, we are careful to hold the leash parallel to the ground to ensure that the human does not provide any upward force that might aid the robot, and our sole purpose is to keep the robot within a safe radius.

\begin{table}[H]
\centering
\resizebox{0.7\columnwidth}{!}{
  \begin{tabular}{l | c c}
  \toprule
  Property & Valid Range & RAPP Search Range \\
  \midrule
  \texttt{robot friction} & $[0, \infty)$ & $[0, 10]$ \\
  \texttt{robot restitution} & $[0, 1]$ & $[0, 1]$ \\
  \texttt{robot payload mass} & $(-\infty, \infty)$ & $[-10, 10]$ \\
  \texttt{robot center of mass displacement} & $(-\infty, \infty)$ & $[-10, 10]$ \\
  \texttt{robot motor strength} & $[0, \infty)$ & $[0, 2]$ \\
  \texttt{robot motor offset} & $(-\infty, \infty)$ & $[-10, 10]$ \\
  \midrule
  \texttt{ball mass} & $[0, \infty)$ & $[0, 10]$ \\
  \texttt{ball friction} & $[0, \infty)$ & $[0, 10]$ \\
  \texttt{ball restitution} & $[0, 1]$ & $[0, 1]$ \\
  \texttt{ball drag} & $[0, \infty)$ & $[0, 10]$ \\
  \midrule
  \texttt{terrain friction} & $[0, \infty)$ & $[0, 10]$ \\
  \texttt{terrain restitution} & $[0, 1]$ & $[0, 1]$ \\
  \texttt{terrain roughness} & $[0, \infty)$ & $[0, 10]$ \\
  \midrule
  \texttt{robot push velocity} & $[0, \infty)$ & $[0, 10]$ \\
  \texttt{ball push velocity} & $[0, \infty)$ & $[0, 10]$ \\
  \texttt{gravity} & $(-\infty, \infty)$ & $[-10, 10]$ \\
  \bottomrule
 \end{tabular}}
  \caption{\textbf{Domain randomization parameters for globe walking, along with their valid ranges and RAPP search ranges.}}
  \label{table:globe_walking_dr_parameters}
\end{table}

\subsection*{E. Additional Ablation Results}
\label{appendix:reward-ablation}
\begin{table}
\centering
\resizebox{0.7\linewidth}{!}{
  \begin{tabular}{l|cc}
  \toprule
  {Sim-to-real Configuration} & Forward Velocity (m/s) & Meters Traveled (m) \\
  \midrule
  Our Method (Average) & 1.66 $\pm$ 0.25 & 4.64 $\pm$ 0.78 \\ 
  Without DR & 1.21 $\pm$ 0.39 & 4.17 $\pm$ 1.04 \\
  With \texttt{Human-Designed} DR & 1.35 $\pm$ 0.16 & 4.83 $\pm$ 0.29 \\
  With Prompt DR & 1.43 $\pm$ 0.45 & 4.33 $\pm$ 0.58 \\
  Without Prior & 0.09 $\pm$ 0.36\footnotemark[1] & 0.31 $\pm$ 1.25 \\
  With Uninformative Prior & 0.08 $\pm$ 0.33\footnotemark[1] & 0.28 $\pm$ 1.13 \\
  With Random Sampling & 0.98 $\pm$ 0.45 & 2.81 $\pm$ 1.80 \\
  \bottomrule
 \end{tabular}}
   \caption{\textbf{Ablations result.} Ablations of the DR formulation in \ourmethod all result in decreased performance.}
  \label{table:main-ablations}
\end{table}

\begin{figure}
\centering
\includegraphics[width=0.45\columnwidth]{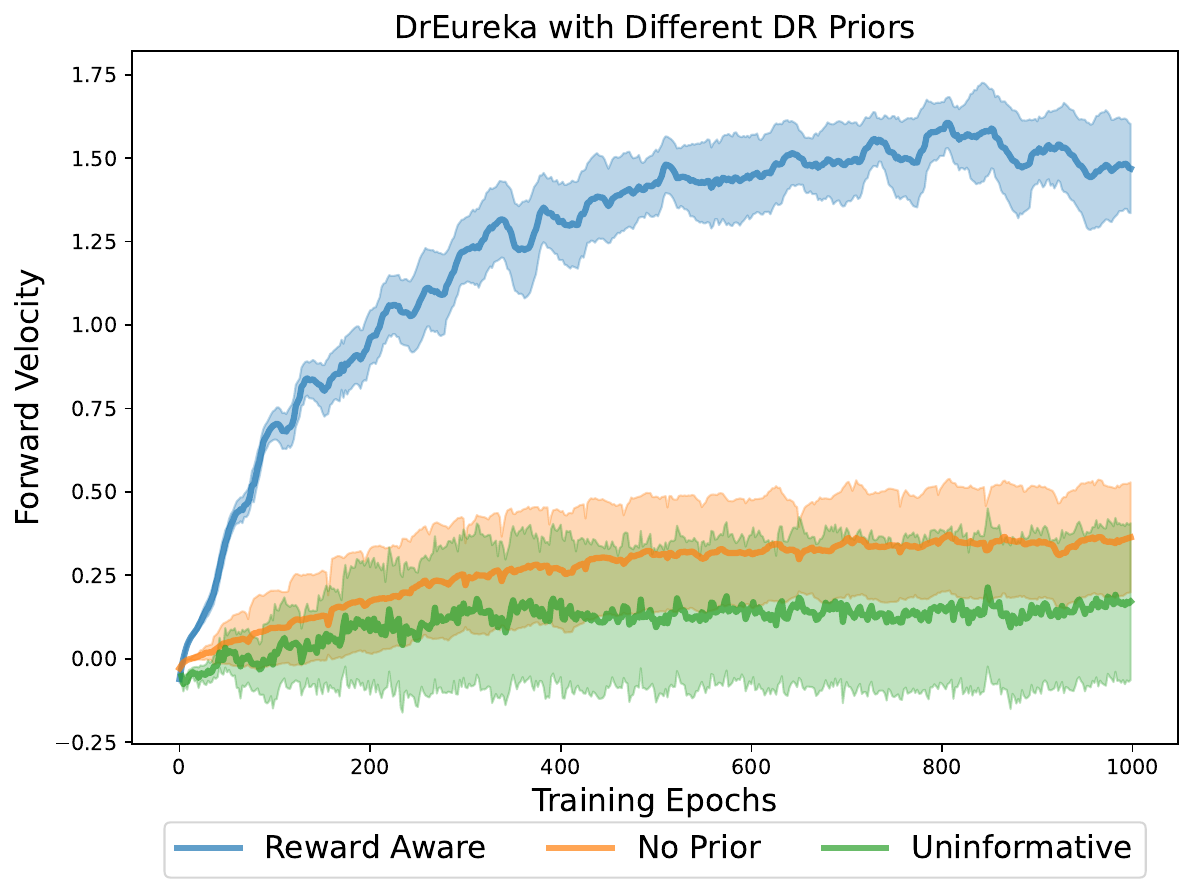}
\caption{Ablations for different domain randomization priors. Replacing RAPP with other choices makes the LLM generate configurations that are difficult to train in simulation.}
\label{figure:dr-comparison}
\end{figure}

\textbf{Sampling from \ourmethod priors enables stable simulation training.} To better understand the drastically different performances of different \ourmethod prior choices in the real world, we present the simulation training curves in Figure~\ref{figure:dr-comparison}. Note that the performances are not directly comparable as each method is trained and evaluated on its own DR distributions. Nevertheless, we observe the stable training progress of \ourmethod. In contrast, despite using a LLM, the ablations synthesize poor DR ranges, resulting in difficult policy training dynamics.

\begin{figure*}[ht]
    \centering
    \includegraphics[width=\linewidth]{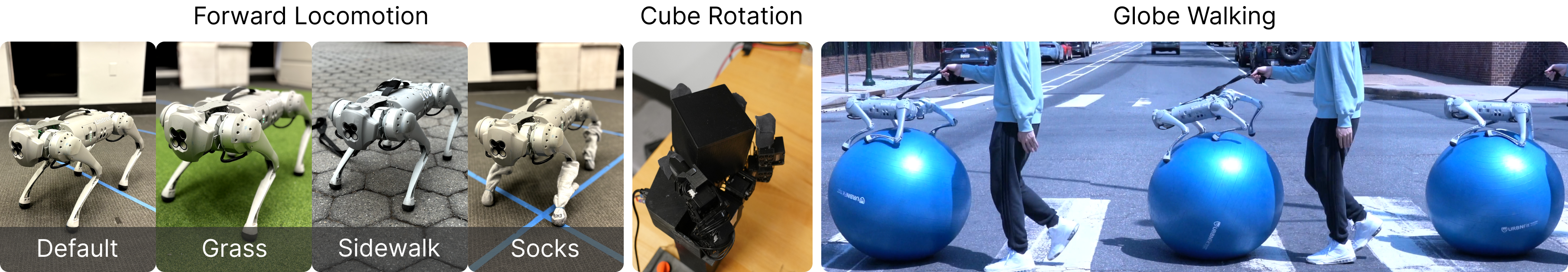}
    \caption{Our forward locomotion, cube rotation, and globe walking tasks.}
    \label{figure:tasks}
\end{figure*}

\begin{table}
\centering
\resizebox{0.7\columnwidth}{!}{
  \begin{tabular}{l|ccc}
  \toprule
  {Safety Instruction} & Velocity (Sim) & Velocity (Real) \\
  \midrule  %
    Yes (\ourmethod w.o DR) & 1.70 $\pm$ 0.11 & \textbf{1.21} $\pm$ 0.39 \\ 
    No  (Eureka) & \textbf{1.83} $\pm$ 0.05 & 0.0 $\pm$ 0.0  \\ 
  \bottomrule
 \end{tabular}}
  \caption{\textbf{\ourmethod safety instruction ablation.} Omitting the safety instruction from \ourmethod results in policies that run quickly in simulation but fail in the real world.}
  \label{table:safety-ablation}
\end{table}

\begin{figure}
\centering
\includegraphics[width=0.45\columnwidth]{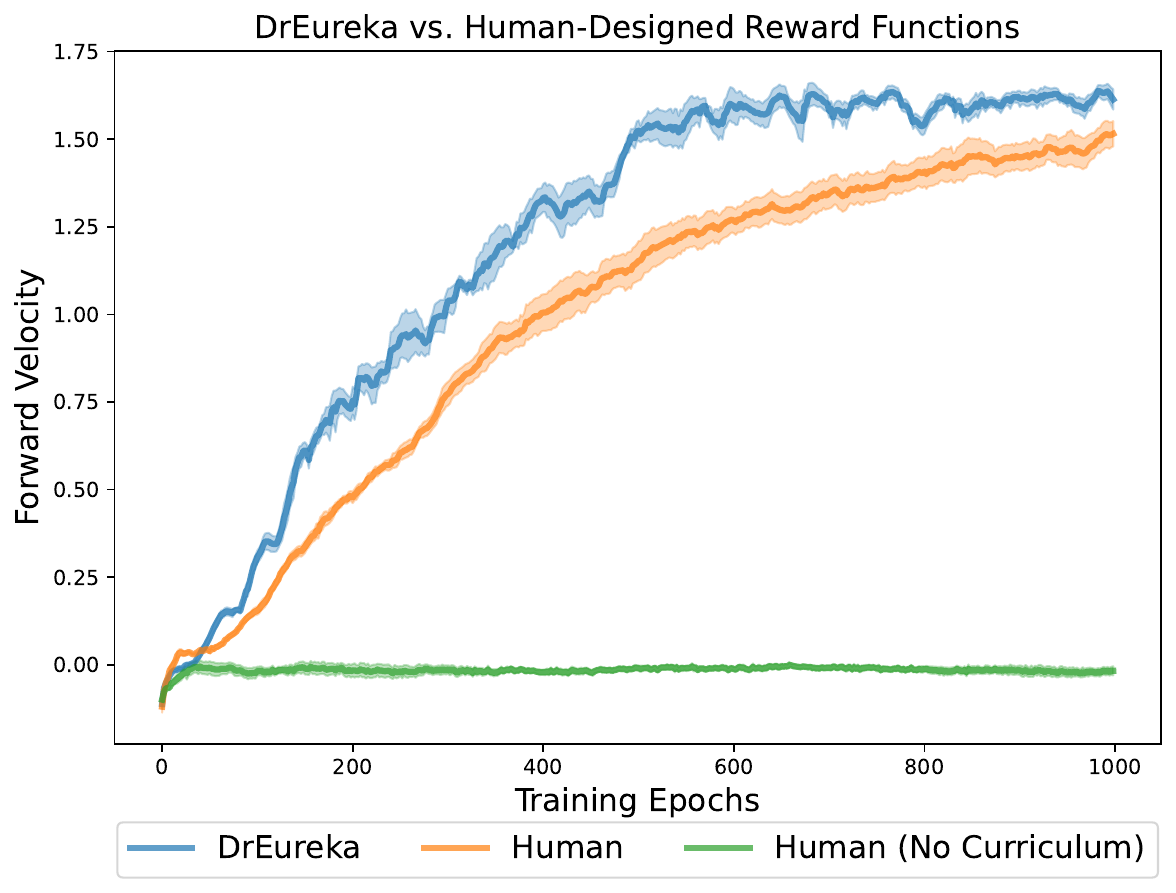}
\caption{Comparison between \ourmethod and \texttt{Human-Designed} reward functions on the simulation locomotion task. \ourmethod has higher sample efficiency and asymptotic performance, while \texttt{Human-Designed} relies on a velocity curriculum to perform well.}
\label{figure:reward-comparison}
\end{figure}

\textbf{\ourmethod does not need a reward curriculum.} To study the effectiveness of the reward functions in isolation, we fix the domain randomization configurations to be \texttt{Human-Designed} for both \ourmethod and \texttt{Human-Designed} reward functions and re-train several policies in simulation. Since \texttt{Human-Designed} reward utilizes a velocity curriculum, we also evaluate an ablation of the \texttt{Human-Designed} reward function that has a fixed velocity target (i.e., 2.0 m/s) to put it on an equal footing with the Eureka reward function as a standalone reward function. The training curves are shown in Figure~\ref{figure:reward-comparison}. As shown, 
\ourmethod reward enables more sample-efficient training and reaches higher asymptotic performance. In contrast, the \texttt{Human-Designed} reward crucially depends on the explicit curriculum to work comparably; as a stand-alone reward function without curriculum inputs, \texttt{Human-Designed} makes little progress.

\begin{figure}[h] %
\centering

\includegraphics[width=0.45\columnwidth]{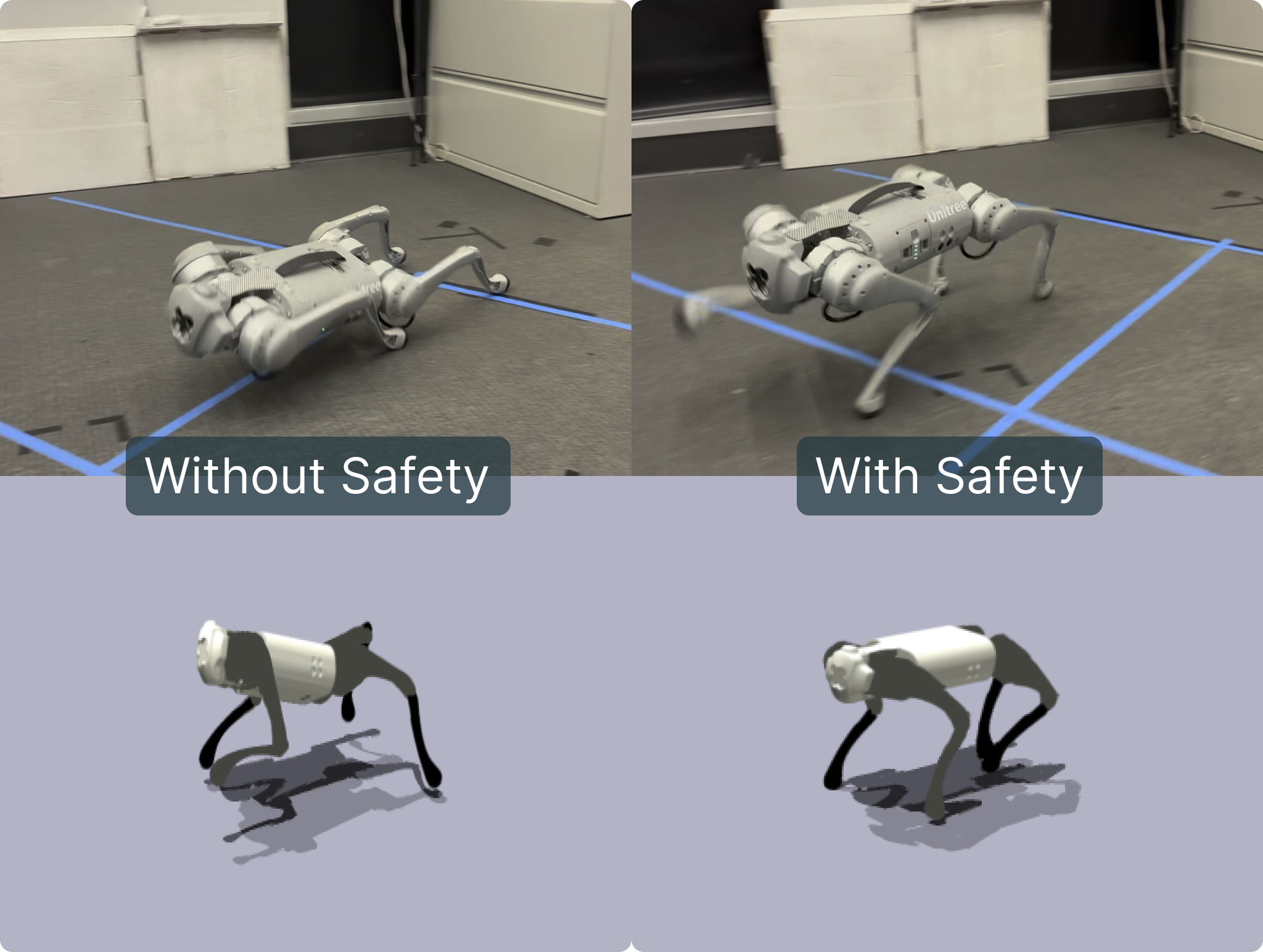}
\caption{\ourmethod with safety instruction successfully learns transferable gait from simulation to real. In contrast, removing the safety instruction leads to behavior that exploits the simulation and quickly fails in the real world.}
\label{figure:safety-ablation}
\end{figure}

\textbf{Safety instruction enables safe reward functions.} In addition to comparing against human-written reward functions, we also ablate \ourmethod's own reward design procedure. In particular, to verify that \ourmethod's safety instruction yields more deployable reward functions, we compare to an ablation of \ourmethod that does not include custom safety suggestions in the prompt; see Table \ref{table:human-written-reward-function-forwardloco} for the functional form of this reward function. Note that this ablation is identical to the original Eureka algorithm in Table~\ref{table:main-results}, and we compare it to the \ourmethod (No DR) variant to eliminate the influence of domain randomization in policy performance. As shown in Table~\ref{table:safety-ablation}, removing the safety prompt results in a final reward function that can move faster in simulation than \ourmethod. However, the robot acquires an unnatural gait with three of its feet and the hip dragging on the ground. Consequently, in the real world, this behavior does not transfer, and the policy directly face-plants at the starting line; this is not surprising as the Eureka reward function contains just a generic action smoothing term for safety, which in itself does not prohibit awkward behaviors. Example snapshots are included in Figure~\ref{figure:safety-ablation}; See our project website for a video comparison.

\subsection*{\newtext{F. CEM and BayRn Baseline Details}}
\newtext{
In this section, we detail our DR baseline procedure for Cross Entropy Method (CEM)~\citep{vuong2019pick, KROESE201319, CrossEntropy} and Bayesian Optimization (BayRn)~\citep{muratore2021data, frazier2018tutorial}. On a high level, both algorithms optimize DR parameters by repeatedly training and evaluating policies in real. Over multiple iterations, CEM trains policies on DR configurations $\mathcal{T}_i, \ldots, \mathcal{T}_j$ sampled from distribution $p$, evaluates their real-world performance $J_i, \ldots, J_j$, and updates $p$ to fit the $k$ "elite" samples with highest $J$. BayRn initially trains and evaluates multiple sampled DR configurations $\mathcal{T}_0, \ldots, \mathcal{T}_i$, then fits a surrogate model $G$ on $(\mathcal{T}_0, J_0), \ldots, (\mathcal{T}_i, J_i)$; next, for multiple iterations, BayRn uses $G$ and acquisition function $a$ to select the next DR configuration $\mathcal{T}_j$ to train and evaluate, then updates $G$ with $(\mathcal{T}_j, J_j)$.
}

\newtext{
For BayRn, we select the widely used Matérn 2.5 kernel and the Upper Confidence Bound (UCB) as the acquisition function with parameter $\kappa = 5$ and $\xi=1$. To maintain the same sample complexity of 16, we run CEM for 4 iterations with 4 samples each and BayRn with 8 initial samples, then 8 iterations with 1 sample each.
}

\end{document}